\documentclass{article}

\PassOptionsToPackage{numbers, compress}{natbib}

\usepackage[preprint]{neurips_2026}

\usepackage[utf8]{inputenc} 
\usepackage[T1]{fontenc}    
\usepackage{algorithm}
\usepackage{algpseudocode}
\usepackage{hyperref}       
\usepackage{url}            
\usepackage{booktabs}       
\usepackage{amsfonts}       
\usepackage{nicefrac}       
\usepackage{microtype}      
\usepackage{xcolor}         
\usepackage{fontawesome5}
\definecolor{darkblue}{rgb}{0.15,0.15,0.48}
\usepackage{amsmath}
\usepackage{amsmath, amssymb, amsthm}
\usepackage{bm}
\usepackage{xspace}
\usepackage{caption}
\usepackage{url} 
\usepackage{booktabs}
\IfFileExists{multirow.sty}{\usepackage{multirow}}{}
\providecommand{\multirow}[3]{#3}
\usepackage[T1]{fontenc}
\usepackage{amsmath}
\usepackage{graphicx}
\usepackage{tikz}
\usepackage{enumitem}
\usepackage{placeins}
\usepackage{subcaption}
\usepackage{pgfplots}
\usepackage{tcolorbox}
\usepackage{enumitem}
\usepackage{tabularx}
\pgfplotsset{compat=1.18} 
\usepackage{listings}
\usepackage[table]{xcolor}
\usepackage{tabularx}
\definecolor{skillbg}{RGB}{249,246,234}
\definecolor{skilltitle}{RGB}{208,207,228}
\definecolor{skillsection}{RGB}{235,235,235}
\definecolor{skillalt}{RGB}{252,250,243}    

\newcolumntype{L}[1]{>{\raggedright\arraybackslash}p{#1}}
\newcolumntype{Y}{>{\raggedright\arraybackslash}X}

\title{\ours{}: Self-Evolving Skill Harnesses for Image Generation Workflows}

\author{\textbf{Zongxia Li}\textsuperscript{1}\thanks{Equal contribution.} \quad \textbf{Dawei Liu}\textsuperscript{2}\footnotemark[1] \quad \textbf{Fuxiao Liu}\textsuperscript{3} \quad \textbf{Yuhang Zhou}\textsuperscript{1} \quad \textbf{Xiyang Wu}\textsuperscript{1}\\ \textbf{Jingxi Chen}\textsuperscript{1} \quad \textbf{Jing Xie}\textsuperscript{1} \quad \textbf{Xiaomin Wu}\textsuperscript{1} \quad \textbf{Lichao Sun}\textsuperscript{4}\\[0.5ex] \textsuperscript{1}University of Maryland \quad \textsuperscript{2}University of Pennsylvania  \quad \textsuperscript{3}Nvidia \quad \textsuperscript{4}Lehigh University \\ {\tt\small zli12321@umd.edu \quad liudawei@seas.upenn.edu \quad lis221@lehigh.edu} \\[0.4ex] \makebox[\textwidth][c]{\small
  \href{https://github.com/Moms-Organic-Agent-Lab/comfyclaw}{\textcolor{darkblue}{\faIcon{github}}}\kern0.35em
  \textbf{Code:}\kern0.35em
  \url{https://github.com/Moms-Organic-Agent-Lab/comfyclaw}} }

\providecommand{\ours}{\textsc{ComfyClaw}\xspace}
\providecommand{\baseline}{\textsc{Base}\xspace}
\providecommand{\comfygems}{\textsc{ComfyGEMS}\xspace}
\providecommand{\skillset}{\Sigma}

\begin{document}

\maketitle

\begin{abstract}

Agents are increasingly used to construct workflows and help humans complete recurring tasks more efficiently. As these workflows become repeated and domain-specific, agent memory and reusable skills become increasingly important: agents should be able to recall workflow patterns, execution constraints, and user preferences from past runs.
We study this problem in workflow-based image generation and introduce \ours{}, an agentic skill evolution harness for controlling ComfyUI workflows. 
\ours{} represents workflow construction as typed graph editing, exposes tools organized by construction stage, reverts back invalid edits, and uses a region-level vision-language model (VLM) verifier to translate visual failures into actionable repair suggestions. 
The framework further evolves a progressively disclosed skill library, where trajectories, execution errors, and verifier feedback from previous runs are distilled into reusable \texttt{Agent~Skills}.
Across four benchmark splits, three agent models, and two image backbones, \ours{} achieves the best average image-generation evaluation score across all six agent configurations, outperforming a verifier-only baseline without skill evolution. 
Human annotations further show that annotators prefer \ours{} over variants without skill evolution. 
Our results suggest that skill evolution is an effective mechanism for improving agent reliability and performance in recurring visual workflow construction.
%

\end{abstract}

\section{Introduction}
\label{sec:introduction}

Agents are moving beyond prompt-only chat interfaces toward workflow execution~\cite{cao2025controllable,xu2025comfyui}. 
This shift is especially important for image generation, where prompt-only interfaces obscure many of the decisions that shape the final result, such as how conditioning is applied, how visual components are composed, and how generated image failures are detected and repaired.
Workflow systems such as ComfyUI~\cite{comfyui_registry} expose these decisions as editable pipelines, making image generation more inspectable, controllable, and reusable~\cite{gal2024comfygen,huang2025comfygpt}. 
At the same time, this fine-grained control turns image creation into a workflow problem: an agent must select compatible components, satisfy model constraints, diagnose visual failures, and repair the workflow without breaking execution. 
Because similar workflow patterns recur across image-generation tasks, effective agents must do more than plan and reflect within a single run: they should also acquire reusable skills from prior executions to avoid repeating the same errors~\cite{bai2026skilldag,liu2026graph}.

This shift changes the agent's role from specifying prompts to operating an executable procedure.
Each step is mediated by a harness that exposes graph edits, runtime feedback, and recovery mechanisms; one invalid operation can break execution, and some failures are only revealed after the workflow is run~\cite{wang2023voyager,wu2024copilot,xie2024osworld,workarena2024,chezelles2025browsergym,xu2025agents,gal2024comfygen,pan2026natural,su2026comfysearch}.
Recent workflow-generation agents reduce this burden by generating or refining workflows from natural language~\cite{xu2025comfyui,jiang2026genagent,xue2025comfybench}, while frameworks such as GEMS show promises of closed-loop refinement, memory, and skills for multimodal generation~\cite{he2026gems}.
However, their control interfaces and skill libraries are static, or updated only by passively storing experience as memory rather than actively refining it into reusable skills.
Thus, a refinement that succeeds in one run rarely becomes a validated procedure that can be invoked safely in future workflows~\cite{wang2023voyager}.

This limitation motivates us that workflow agents need infrastructure for both controlling the current executable state and carrying useful experience for future tasks.
Turning a workflow repair into reusable workflow competence requires two components: a \emph{harness} that exposes tools, feedback, memory, and state transitions to the agent~\cite{madaan2023self,shinn2023reflexion,young2025effective,lou2026autoharness,lee2026meta,xu2026agent}, and \emph{skill management}, which converts repeated trajectories and past execution experiences into reusable procedural knowledge~\cite{wang2023voyager,wu2025evolver,xia2026skillrl,ni2026trace2skill,zhai2025agentevolver,gao2025survey,fang2025comprehensive}.
A harness without evolving skills repeatedly rediscovers similar workflow repairs, while skills without a strong harness can become brittle instructions that cause errors for the current workflow.
Thus, the challenge is not only to refine a workflow during one run, but to turn feedback from that run into reusable control knowledge for future runs.
This motivates our question: \emph{can feedback from an agent's self-verifier support both immediate workflow repair and the long-term evolution of a reusable skill library?}

Inspired by prior work on agentic workflow control and self-evolving agents~\cite{wu2025evolver,zhai2025agentevolver,gal2024comfygen,sobania2024comfygi}, we present \ours{}, a self-evolving framework for controlling image-generation workflows in an \emph{unmodified} ComfyUI runtime.
We cast workflow execution as a skill-augmented Markov Decision Process over executable graph edits, runtime feedback, verifier feedback, and reusable skills, making harness design and skill reuse explicit parts of the control problem.
Built on this formulation, \ours{} combines typed graph editing for workflow construction, VLM-based verification for diagnosing and repairing visual failures, and a skill-evolution loop that proposes, validates, and commits reusable Agent~Skills.
We evaluate \ours{} across four benchmark splits, two image backbones, and three agent models, and find that it achieves the best average score, outperforming harness-only control by 4 absolute points and no-refinement control by 10 absolute points.
Human annotators further prefer \ours{} on 2{,}400 images, and 318 evolved skills account for roughly $50\%$ of later skill invocations.

\section{Related Work}

\textbf{Workflow graphs as controllable creative artifacts.}
Workflow graphs are common in creative software, from Blender nodes~\cite{blender} and Houdini networks~\cite{houdini} to Nuke compositing~\cite{nuke} and Unreal Blueprints~\cite{unrealengine}. In these systems, the graph is the artifact: it exposes intermediate structure, supports reuse, and makes complex pipelines easier to inspect than monolithic code. ComfyUI~\cite{comfyanonymous2023comfyui} brings the same idea to diffusion-based generation, where users build image, video, and audio-visual pipelines from node graphs~\cite{xu2025comfyui,guo2025comfymind}. Effective use, however, still depends on knowledge of node compatibility, model constraints, and scattered community recipes~\cite{angert2023spellburst,tamilselvam2019visual,vuruma2024cloud,huang2025comfygpt,guo2025comfymind}. Recent work therefore treats ComfyUI as an agent-control target: ComfyGen~\cite{gal2024comfygen} selects workflows from prompts, ComfyGPT~\cite{huang2025comfygpt} and GenAgent~\cite{jiang2026genagent} synthesize graphs or code through multi-agent collaboration, ComfyUI-R1~\cite{xu2025comfyui} studies RL-tuned reasoning, and ComfyBench~\cite{xue2025comfybench} provides an evaluation setting. These systems mostly optimize workflow generation or refinement for a single prompt. In contrast, \ours{} treats workflow construction as closed-loop control: it edits executable graphs through a typed, stage-gated harness, repairs failures with localized verifier feedback, and promotes reusable skills only after held-out validation under a graph-complexity prior.

\textbf{Harnessed agents with reusable skills.}
Recent agent systems use explicit skills, tool interfaces, and runtime harnesses to make long-horizon execution more reliable. Voyager~\cite{wang2023voyager} set an influential template, where an LLM agent improves through an automatic curriculum, executable skill library, and iterative self-verification. Later work developed related pieces, including reward design in Eureka~\cite{ma2023eureka}, critique and reflection in Self-Refine~\cite{madaan2023self} and Reflexion~\cite{shinn2023reflexion}, desktop-task scaffolding in OS-Copilot~\cite{wu2024copilot}, and declarative pipeline construction in DSPy~\cite{khattab2023dspy}. A parallel line treats skills as reusable agent artifacts. The \textit{Anthropic Agent Skills} specification~\cite{agentskills2026specification} defines a lightweight \texttt{SKILL.md} format with progressive disclosure, used by systems such as Claude Code~\cite{anthropic2025claudecode}, Hermes Agent~\cite{nousresearch2026hermesagent}, and OpenClaw~\cite{openclaw2026}. Recent methods also learn or revise skills from experience: SkillRL~\cite{xia2026skillrl} builds a hierarchical SkillBank, EvoSkill~\cite{alzubi2026evoskill} mines and repairs skills from failures with held-out validation, and COS-PLAY~\cite{wu2026coevolvingllmdecisionskill} co-evolves decision and skill-bank agents from unlabeled game rollouts.
\ours{} brings this idea to workflow control: it combines typed, stage-gated graph editing with localized verifier feedback and held-out validation, so workflow skills can be learned, tested, and reused rather than manually written or statically retrieved.

\textbf{Skill-centric agents and self-improving harnesses.}
LLM agents are increasingly designed to act on a user's behalf over long horizons, often through harnesses that combine tool use, memory, delegation, and reusable skills. Recent open-source systems illustrate this shift: DeerFlow 2.0~\cite{bytedance2025deerflow} uses a LangGraph-based harness with sandboxed execution, persistent memory, sub-agents, and extensible skills; OpenClaw~\cite{openclaw2026} builds a multi-channel assistant around skill-based operation; and Hermes Agent~\cite{nousresearch2026hermesagent} emphasizes reusable skills and autonomous skill creation. A related research line studies skill accumulation and self-improvement as learning objectives, including XSkill~\cite{jiang2026xskill}, EvolveR~\cite{wu2025evolver}, and broader surveys of self-evolving agents~\cite{gao2025survey,fang2025comprehensive}. Other systems target human-facing long-horizon work more directly: Odysseus~\cite{shi2026odysseusscalingvlms100} studies stable RL for vision-language models (VLM) agents in long-horizon visual control, while AgentLab~\cite{schmidgall2025agent} executes research workflows with human feedback. 
These systems show the promise of skill-centric agents, but their skills are often used in broad, open-ended settings where effectiveness and reuse are hard to evaluate. 
\section{Method}
\label{sec:method}

\begin{figure}
    \centering
    \includegraphics[width=0.7\linewidth]{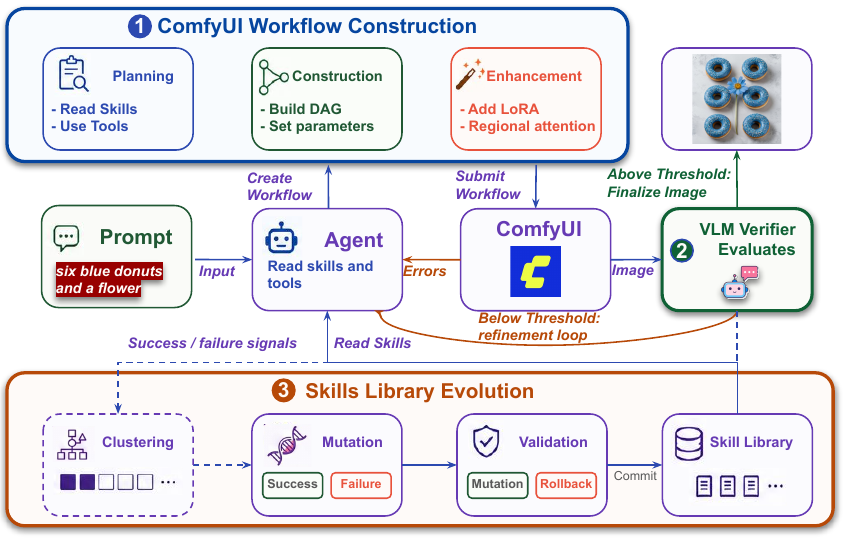}
    \caption{\textbf{Overall framework of \ours{}.}
    The agent edits a ComfyUI workflow graph, the runtime renders a candidate image, the verifier returns requirement-level and region-level feedback, and the agent evolves skills that can be reused in future workflow-construction runs.}
    \label{fig:overview}
    \vspace{-10pt}
\end{figure}


\ours{} is an agentic framework for workflow-based image generation with reusable workflow skills. It has three components: \emph{workflow construction}, \emph{verifier-guided refinement}, and \emph{skill evolution}. 
Given a prompt $p$, an LLM agent constructs a ComfyUI workflow graph through typed edits and submits it to the runtime for rendering. 
A VLM verifier scores the image against the prompt and returns localized repair feedback. 
The agent uses this feedback, along with runtime errors, to refine the workflow over multiple iterations.
Across prompts, recurring successes and failures are distilled into validated Agent Skills, stored in a skill library, and retrieved in future runs.
Figure~\ref{fig:overview} summarizes the pipeline.




\subsection{Preliminaries: Workflows and Agent Skills}




\textbf{ComfyUI workflow graphs.}
We represent a ComfyUI workflow as a directed graph $G=(V,E)$.
Nodes $V$ define the operations in the generation pipeline, including text encoding, latent sampling, LoRA loading, regional conditioning, upsampling, inpainting, and image decoding.
Edges $E$ carry intermediate outputs from one operation to the next.
Given a prompt $p$, ComfyUI executes the graph to render an image or video.
This graph-based interface gives the agent more control than prompt-only generation, since it can revise both the prompt and the pipeline that processes it.




\textbf{Workflow editing.}
We treat workflow construction as a sequence of graph edits.
The agent can add or remove nodes, connect or disconnect edges, and adjust node parameters. Once constructed, the workflow is submitted to ComfyUI for execution.


\textbf{Agent skills.}
Agent skills are reusable procedures stored as \texttt{SKILL.md} files.
Each skill contains a brief description, optional triggers, and step-by-step instructions for editing a workflow graph.
To keep the context small, the agent is initially shown only lightweight skill metadata, such as the skill name and description, and retrieves the full instructions only when it decides that a skill is relevant.
This progressive-disclosure design lets skills serve as retrievable workflow-editing knowledge that can be reused across runs and revised after completed editing attempts.


\subsection{Workflow Construction}
\label{sec:method:workflow-construction}
\noindent\textbf{Planning and initialization.}
Given a prompt $p$, the agent first enters a planning stage, where it identifies the target image-generation model, reviews relevant tools and skills, and decides how to construct the workflow.
We represent the workflow as a directed acyclic graph (DAG),
$G_t = (V_t, E_t)$,
where $V_t$ denotes the set of workflow nodes and $E_t$ denotes the data-flow edges at refinement step $t$.
The agent begins by constructing an initial workflow $G_0$, a minimal \emph{spine} graph that loads the diffusion model, encodes the prompt, samples the latent representation, decodes the image, and saves the output.
This initial graph provides a valid starting point for later refinement.

\noindent\textbf{Graph editing.}
The agent refines the workflow through a sequence of graph-editing actions,
\begin{equation}
    G_{t+1} = \mathrm{Edit}(G_t, a_t),
    \qquad a_t \in \mathcal{A},
\end{equation}
where $\mathcal{A}$ includes node insertion, node connection, parameter updates, prompt edits, LoRA insertion, regional conditioning, and refinement passes.
The objective of workflow construction is to produce a final workflow $G_T$ that reaches the verifier reward threshold (Section~\ref{subsec:vlm-verifier}).

\noindent\textbf{Skill retrieval.}
Skills are available throughout workflow construction but are exposed selectively through a trigger-based router.
Each skill $\sigma_i \in \skillset$ includes lightweight metadata, such as a name, short description, tags, and trigger phrases.
Given the current workflow stage and verifier feedback, the agent scores each skill by its relevance to this metadata, and only the top-$K$ skills are shown initially.
The full skill content is retrieved only when needed.
This progressive exposure keeps the context small while still allowing access to detailed procedures for workflow construction and repair.

\noindent\textbf{Construction stages.}
After creating the spine graph, the agent enters the construction stage, where it edits the workflow DAG by adding or connecting nodes, setting parameters, and modifying the graph structure to satisfy the prompt requirements.
It then proceeds to an enhancement stage, where it can apply higher-level workflow changes such as adding LoRA modules, regional attention, mask-based conditioning, or refinement passes.
Together, these stages correspond to the planning, construction, and enhancement blocks shown at the top of Figure~\ref{fig:overview}.




\subsection{Verifier-Guided Refinement}
\label{subsec:vlm-verifier}
After the agent submits $G_t$, ComfyUI renders an image $I_t$ which is
passed to a VLM verifier.
The verifier first decomposes $p$ into a checklist of observable binary requirements $Q=\{q_i\}$.
For each $q_i$, it returns a binary pass/fail label grounded in the
image, together with a short natural-language justification; in
addition it returns a holistic detail score $s_{\mathrm{det}}\!\in\![0,10]$
that captures overall fidelity, composition, and absence of visual artifacts.
The harness combines the two signals into the scalar reward of
Eq.\,\ref{eq:reward}.

\paragraph{Vision-Language Model (VLM) as a Verifier.}
We use the agent as verifier to evaluate whether a generated image satisfies the prompt.
The verifier decomposes the prompt into observable requirements
$Q=\{q_i\}$, such as object count, attribute binding, spatial relation, style,
and anatomy.
It returns requirement-level pass or fail labels, a holistic detail score, localized failure descriptions, and concrete suggestions for workflow edits.
The score used by the harness is
\begin{equation}
  r \;=\; 0.6\cdot\!\frac{|Q^{\mathrm{pass}}|}{|Q|}
        \;+\; 0.4\cdot\!\frac{s_{\mathrm{det}}}{10},
  \label{eq:reward}
\end{equation}
where $s_{\mathrm{det}}$ is an additional one-pass quality score, ranging from 1 to 10, that measures the overall image quality and its alignment with the prompt.

\paragraph{Workflow Refinement loop.}
The verifier emits three structured components that drive the next iteration: (i) the failing-requirement set
$Q\!\setminus\!Q^{\mathrm{pass}}$, (ii) localized natural-language
descriptions of what went wrong (e.g.\ \textit{the leftmost figure has three arms}), and (iii) concrete edit suggestions phrased in workflow terms (\textit{apply regional prompting to isolate the throwing arm}).
These pieces are appended to the agent's next-iteration context and also forwarded to the skill-evolution loop (\S\ref{sec:skill-evolution}).
The refinement loop terminates when $r$ exceeds a satisfaction threshold
$\tau_{\mathrm{stop}}$ or after $K$ iterations, whichever comes first;
the iteration with the highest $r$ is committed as the final output.


\subsection{Skill Evolution}
\label{sec:skill-evolution}
The skill library serves as the long-term memory of \ours{}. 
Rather than treating each prompt independently, \ours{} converts recurring successes and failures across prompts into reusable workflow procedures. 
Skill evolution proceeds in four stages: clustering success and failure traces, proposing skill mutations, validating the proposed mutations, and committing accepted skills to the skill library.

\textbf{Success and Failure Clustering.}
Let $\mathcal{S}^{(m)}$ denote the skill library after evolution cycle $m$.
After a batch of $B$ prompts, the workflow-construction loop produces traces
\begin{equation}
    \mathcal{D}^{(m)}
    =
    \left\{
    \left(p_i, G_i, x_i, r_i, f_i, e_i, a_{i,1:T_i}\right)
    \right\}_{i=1}^{B},
\end{equation}
where $p_i$ is the prompt, $G_i$ is the final workflow, $x_i$ is the generated
image, $r_i$ is the verifier reward, $f_i$ is verifier feedback, $e_i$ is a
runtime error if one occurs, and $a_{i,1:T_i}$ is the sequence of workflow
actions.
As shown in the bottom part of Figure~\ref{fig:overview}, these success and
failure traces are sent to the skill evolution module.

We divide traces into success and failure groups using the verifier reward.
A trace is treated as successful if $r_i \ge 0.9$ and as a failure otherwise.
We then cluster the success and failure traces separately according to their
verifier feedback, runtime errors, workflow actions, and prompt properties.
Failure clusters capture recurring failure modes, such as missing objects,
incorrect counting, and weak spatial binding.
Success clusters capture reusable workflow strategies, such as when to add a LoRA, when to use regional attention, how to adjust guidance, or how to structure prompts.

\textbf{Skill Mutation.}
For each cluster $C_\ell$, we use the agent as a skill evolver to propose a
mutation to the current skill library.
The mutation type is selected from
\begin{equation}
    \mu_\ell \in
    \{\texttt{create}, \texttt{revise}, \texttt{reinforce},
      \texttt{merge}, \texttt{delete}\}.
\end{equation}
A \texttt{create} mutation adds a new procedure for a pattern not covered by
existing skills.
A \texttt{revise} mutation updates the content of an existing skill.
A \texttt{reinforce} mutation strengthens useful triggers or emphasis terms.
A \texttt{merge} mutation combines redundant skills, and a \texttt{delete}
mutation removes skills that are no longer useful.
Applying the proposed mutation to the current library gives a candidate skill
library
\begin{equation}
    \widetilde{\mathcal{S}}_{\ell}^{(m+1)}
    =
    \mu_\ell\!\left(\mathcal{S}^{(m)}\right).
\end{equation}

\textbf{Mutation Validation.}
Each candidate mutation is validated before it is committed to the skill
library.
For cluster $C_\ell$, we ask the agent to synthesize three held-out prompts
conditioned on the cluster identifier, the candidate skill name, and up to five
example prompts from the cluster.
These prompts are intended to test whether the proposed mutation generalizes
beyond the examples that produced it.
%

Let $\mathcal{H}_\ell$ denote the validation prompt set for cluster $C_\ell$.
We compare the average verifier reward obtained with the current skill library
$\mathcal{S}^{(m)}$ against the reward obtained with the candidate library
$\widetilde{\mathcal{S}}_{\ell}^{(m+1)}$:
\begin{equation}
    \Delta_\ell
    =
    \frac{1}{|\mathcal{H}_\ell|}
    \sum_{p \in \mathcal{H}_\ell}
    \left[
    r\!\left(p;\widetilde{\mathcal{S}}_{\ell}^{(m+1)}\right)
    -
    r\!\left(p;\mathcal{S}^{(m)}\right)
    \right].
\end{equation}
The mutation is accepted only if it does not degrade validation performance $\Delta_\ell \ge 0.$
%
Rejected mutations are rolled back, while accepted mutations are committed as
new skill versions.

\begin{table}[!htbp]
  \centering
  \small
  \resizebox{\textwidth}{!}{%
  \begin{tabular}{lllrrrrr}
    \toprule
    \textbf{Agent} & \textbf{Image Model} & \textbf{Method} & \textbf{GenEval2} $\uparrow$ & \textbf{DPG-Bench} $\uparrow$ & \textbf{OneIG-EN} $\uparrow$ & \textbf{OneIG-ZH} $\uparrow$ & \textbf{Avg} $\uparrow$ \\
    \midrule
    \multirow{6}{*}{\texttt{claude-sonnet-4.5}} & \multirow{3}{*}{\texttt{longcat}} & \baseline{} & 31.31 & 88.65 & 74.65 & 73.72 & 67.08 \\
     &  & \ours{} & \textbf{54.76} & 90.25 & \textbf{83.64} & \textbf{77.36} & \textbf{75.52} \\
    \cmidrule(lr){2-8}
     & \multirow{3}{*}{\texttt{z-image-turbo}} & \baseline{} & 41.88 & 89.00 & 71.56 & 69.30 & 67.94 \\
     &  & \ours{} & \textbf{62.01} & \textbf{91.12} & 80.93 & \textbf{77.06} & \textbf{77.78} \\
    \midrule
    \multirow{6}{*}{\texttt{Qwen-3.6-35B-A3B}} & \multirow{3}{*}{\texttt{longcat}} & \baseline{} & 30.40 & 88.18 & 72.15 & 69.47 & 65.05 \\
     &  & \ours{} & \textbf{47.08} & \textbf{90.97} & \textbf{86.86} & \textbf{80.46} & \textbf{76.34} \\
    \cmidrule(lr){2-8}
     & \multirow{3}{*}{\texttt{z-image-turbo}} & \baseline{} & 36.53 & 88.89 & 66.95 & 62.97 & 63.84 \\
     &  & \ours{} & \textbf{58.24} & \textbf{91.14} & \textbf{84.98} & \textbf{80.14} & \textbf{78.62} \\
    \midrule
    \multirow{6}{*}{\texttt{Gemma-4-E4B-it}} & \multirow{3}{*}{\texttt{longcat}} & \baseline{} & 16.41 & 65.55 & 37.31 & 37.01 & 39.07 \\
     &  & \ours{} & 21.94 & \textbf{71.33} & \textbf{42.35} & \textbf{40.15} & \textbf{43.94} \\
    \cmidrule(lr){2-8}
     & \multirow{3}{*}{\texttt{z-image-turbo}} & \baseline{} & 32.76 & 85.84 & 59.81 & 57.30 & 58.93 \\
     &  & \ours{} & \textbf{36.21} & \textbf{89.70} & 67.27 & \textbf{66.85} & \textbf{65.01} \\
    \bottomrule
  \end{tabular}
  }
  \caption{\textbf{\ours{} achieves the best overall performance across the four benchmarks.} All scores are reported on a 0--100 scale, where higher is better. Scores are judged by Qwen3-VL-8B-Instruct using each benchmark's evaluation metric: Soft-TIFA Geometric Mean for GenEval2, Soft-TIFA Arithmetic Mean for DPG-Bench, and VQAScore for OneIG-EN and OneIG-ZH. The best score in each column is shown in bold.}
  \label{tab:results1}
  \vspace{-10pt}
\end{table}

\section{Experiments}
\label{sec:results}

In this section, we evaluate \ours{} on four text-to-image benchmark splits against competitive baselines. We further present ablation studies and qualitative analyses that illustrate how \ours{} improves through iterative workflow refinement.


\textbf{Benchmarks.}
We evaluate on four text-to-image benchmark splits covering compositional reasoning, dense prompt following, fine-grained fidelity, and cross-lingual generalization.
\textbf{GenEval~2}~\cite{kamath2025geneval} includes 800 English prompts over objects, attributes, and spatial or numerical relations, scored with the atom-level Soft-TIFA VQA judge to avoid saturation in the original GenEval.
\textbf{DPG-Bench}~\cite{hu2024ella} contains 1{,}065 dense English prompts with multiple objects, attributes, and relations in each sentence.
\textbf{OneIG-Bench}~\cite{chang2025oneig} evaluates subject--element alignment, text rendering, reasoning content, stylization, and diversity. We use its English split (\textbf{OneIG-EN}, 1{,}120 prompts) and Chinese split (\textbf{OneIG-ZH}, 1{,}320 prompts).
Together, these benchmarks contain roughly 4{,}300 prompts and test the dimensions where workflow evolution should differ most from prompt-only refinement.


\textbf{Agents and ComfyUI models used.}
We use three foundation models as workflow-control agents: Claude Sonnet
4.5~\cite{anthropic2025claude}, Qwen-3.6-35B-A3B~\cite{yang2025qwen3}, and Gemma-4-E4B-it~\cite{gemma4_2026}. 
All are natively multimodal and are used to instantiate the workflow, verifier, and skill-evolution agents. 
For image generation, we evaluate two ComfyUI backbones: z-image-turbo~\cite{cai2025z}, a 6B text-to-image model, and LongCat-Image~\cite{team2025longcat}, a 6B bilingual Chinese--English model for image generation and editing. 
The agents edit the ComfyUI workflow graph, while the selected backbone renders the final image.

\textbf{Workflow and tool setup.}
To closely simulate how an external agent would control a real ComfyUI deployment, we instantiate a fresh ComfyUI server endpoint for each experimental condition.
Each endpoint runs an unmodified ComfyUI server, while our agent is deployed as an external plug-in that submits jobs and retrieves execution logs through the standard interface.
This design allows our framework to interact with ComfyUI without modifying the application itself, making the setup easy to maintain and naturally compatible with any future ComfyUI updates.

We provide a set of workflow-editing tools for controlling ComfyUI workflows,
together with four predefined visual-quality skills by~\citet{he2026gems}.
During execution, the agent does more than prompt rewriting: it can edit the workflow graph, tune node hyperparameters, and, when supported by the image model, attach and configure LoRAs.
The full list of predefined tools, predefined skills, and model-specific LoRA settings is provided in Appendix~\ref{app:tools-skills}.

\textbf{Image Generation Evaluation Metrics.}
We follow the headline metric defined by each benchmark.
For GenEval2, we report the Soft-TIFA geometric mean~\cite{hu2023tifa} (gm), computed over the per-prompt set of VQA questions, including yes/no and counting questions~\cite{lin2024evaluating}.
For DPG-Bench, we report the Soft-TIFA arithmetic mean (am), computed over each prompt’s corresponding set of evaluation questions.
For OneIG-EN and OneIG-ZH, we use VQAScore, where each image is evaluated with a single query of the form: \textit{Does this image show \textit{<prompt>}? Answer Yes or No.}
All of these metrics rely on a VLM-as-a-judge.
In our experiments, we use Qwen3-VL-8B-Instruct~\cite{bai2025qwen3} as the primary judge model to answer the benchmark evaluation queries.

\subsection{Main Results}
%
%
We compare \ours{} against a baseline that uses only the initial tools and skill set.
It does not receive verifier feedback or perform refinement; instead, it executes the workflow once and directly returns the generated image.
%
%
All experiments and evaluations are run on an RTX PRO 6000 Blackwell GPU with 96 GB of memory. 
We report the results in Table~\ref{tab:results1}.
%
Overall, \ours{} outperforms \baseline{} across all four benchmarks.
These results suggest two main takeaways.
First, adding a VLM verifier and refinement loop helps agents construct better workflows and generate higher-quality images.
Second, skill evolution further improves performance by enabling agents to summarize successes and failures, learn from past errors, and update their skill set for future workflow construction.
%



\begin{figure}[!htbp]
    \centering
    \hspace*{-1.5em}
    \definecolor{basecolor}{HTML}{A8BDD4}      
    \definecolor{evolvedcolor}{HTML}{4A7BB0}   
    \definecolor{evt1}{HTML}{4A7BB0}           
    \definecolor{evt2}{HTML}{1F8A4F}           
    \definecolor{evt3}{HTML}{B87A10}           
    \definecolor{evt4}{HTML}{7AAFD1}           
    \definecolor{evt5}{HTML}{6DB88E}           
    \definecolor{evt6}{HTML}{B8BEC4}           

    \begin{subfigure}[t]{0.48\textwidth}
        \vspace{0pt}
        \centering
        \begin{tikzpicture}
            \begin{axis}[
                ybar,
                bar width=11pt,
                width=\linewidth,
                height=5.6cm,
                enlarge x limits=0.18,
                ylabel={\small \# Skill reads (k)},
                xlabel={\vphantom{Composition (\%)}},
                symbolic x coords={DPG, GenEval2, OneIG-EN, OneIG-ZH},
                xtick=data,
                xticklabel style={font=\footnotesize, text width=1.7cm, align=center},
                ymin=0, ymax=14500,
                ytick={0,3000,6000,9000,12000},
                yticklabels={0, 3, 6, 9, 12},
                yticklabel style={font=\scriptsize},
                scaled y ticks=false,
                axis lines*=left,
                ymajorgrids=true,
                grid style={dotted, gray!50},
                area legend,
                legend style={
                    at={(0.5,-0.22)},
                    anchor=north,
                    legend columns=2,
                    draw=none, fill=none,
                    font=\small,
                    column sep=10pt,
                },
                nodes near coords align={vertical},
            ]
            \addplot[fill=basecolor, draw=none] coordinates {
                (DPG,2078)(GenEval2,5377)(OneIG-EN,3334)(OneIG-ZH,2747)};
            \addplot[fill=evolvedcolor, draw=none] coordinates {
                (DPG,6478)(GenEval2,11391)(OneIG-EN,872)(OneIG-ZH,244)};

            \node[font=\scriptsize\bfseries, text=evolvedcolor, anchor=south]
                at (axis cs:DPG,      6478)  {70.0\%};
            \node[font=\scriptsize\bfseries, text=evolvedcolor, anchor=south]
                at (axis cs:GenEval2, 11391) {56.2\%};
            \node[font=\scriptsize\bfseries, text=evolvedcolor, anchor=south]
                at (axis cs:OneIG-EN, 3334)  {16.3\%};
            \node[font=\scriptsize\bfseries, text=evolvedcolor, anchor=south]
                at (axis cs:OneIG-ZH, 2747)  {7.5\%};

            \legend{Predefined skills, Evolved skills}
            \end{axis}
        \end{tikzpicture}
        \caption{\textbf{Skill-read distribution per benchmark} (full counts in Table~\ref{tab:evolved}).
        Evolved skills account for $\mathbf{70.0\%}$ and $\mathbf{56.2\%}$ of all skill reads on DPG and GenEval2, but only $7.5$--$16.3\%$ on the OneIG splits, indicating benchmark-dependent reliance on agent-evolved knowledge.}
        \label{fig:evolved_reads}
    \end{subfigure}
    \hfill
    \begin{subfigure}[t]{0.48\textwidth}
        \vspace{0pt}
        \centering
        \begin{tikzpicture}
            \begin{axis}[
                xbar stacked,
                bar width=14pt,
                width=\linewidth,
                height=5.6cm,
                enlarge y limits=0.22,
                xlabel={\small Composition of workflow events (\%)},
                symbolic y coords={OneIG-ZH, OneIG-EN, GenEval2, DPG},
                ytick=data,
                yticklabels={%
                    \shortstack[r]{OneIG-ZH\\[-1pt]{\color{black!55}\scriptsize$N{=}9.6\mathrm{k}$}},
                    \shortstack[r]{OneIG-EN\\[-1pt]{\color{black!55}\scriptsize$N{=}8.4\mathrm{k}$}},
                    \shortstack[r]{GenEval2\\[-1pt]{\color{black!55}\scriptsize$N{=}10.9\mathrm{k}$}},
                    \shortstack[r]{DPG\\[-1pt]{\color{black!55}\scriptsize$N{=}6.7\mathrm{k}$}}%
                },
                yticklabel style={font=\footnotesize, align=right},
                xmin=0, xmax=105,
                xtick={0,25,50,75,100},
                xticklabel style={font=\scriptsize},
                axis lines*=left,
                xmajorgrids=true,
                grid style={dotted, gray!50},
                area legend,
                legend style={
                    at={(0.42,-0.22)},
                    anchor=north,
                    legend columns=3,
                    draw=none, fill=none,
                    font=\scriptsize,
                    legend cell align=left,
                    column sep=0.8ex, 
                    row sep=0pt,
                },
            ]
            \addplot[fill=evt1, draw=white, line width=0.5pt] coordinates {
                (42.52,DPG)(38.02,GenEval2)(39.07,OneIG-EN)(38.66,OneIG-ZH)};
            \node[font=\scriptsize, text=white] at (axis cs:21.26,DPG)      {43};
            \node[font=\scriptsize, text=white] at (axis cs:19.01,GenEval2) {38};
            \node[font=\scriptsize, text=white] at (axis cs:19.54,OneIG-EN) {39};
            \node[font=\scriptsize, text=white] at (axis cs:19.33,OneIG-ZH) {39};

            \addplot[fill=evt2, draw=white, line width=0.5pt] coordinates {
                (41.24,DPG)(22.18,GenEval2)(40.63,OneIG-EN)(43.32,OneIG-ZH)};
            \node[font=\scriptsize, text=white] at (axis cs:63.14,DPG)      {41};
            \node[font=\scriptsize, text=white] at (axis cs:49.11,GenEval2) {22};
            \node[font=\scriptsize, text=white] at (axis cs:59.39,OneIG-EN) {41};
            \node[font=\scriptsize, text=white] at (axis cs:60.32,OneIG-ZH) {43};

            \addplot[fill=evt3, draw=white, line width=0.5pt] coordinates {
                (3.18,DPG)(27.17,GenEval2)(5.77,OneIG-EN)(4.16,OneIG-ZH)};
            \node[font=\scriptsize\bfseries, text=white]
                at (axis cs:73.79,GenEval2) {27};

            \addplot[fill=evt4, draw=white, line width=0.5pt] coordinates {
                (8.47,DPG)(4.76,GenEval2)(8.60,OneIG-EN)(8.39,OneIG-ZH)};

            \addplot[fill=evt5, draw=white, line width=0.5pt] coordinates {
                (2.90,DPG)(2.37,GenEval2)(3.91,OneIG-EN)(3.73,OneIG-ZH)};

            \addplot[fill=evt6, draw=white, line width=0.5pt] coordinates {
                (1.69,DPG)(5.50,GenEval2)(2.03,OneIG-EN)(1.74,OneIG-ZH)};

            \legend{Prompt-text, Sampler, Regional,
                    Model/weight, Multi-pass, Other}

            \end{axis}
        \end{tikzpicture}
        \caption{\textbf{Workflow-edit composition per benchmark} (each bar $=100\%$; $N$\,=\,total events per benchmark shown below each axis label; full counts in Table~\ref{tab:modification-breakdown}).
        Non-prompt edits make up $\mathbf{60.7\%}$ of all events; GenEval2 uniquely allocates $\mathbf{27\%}$ of edits to regional/mask topology vs.\ $3$--$6\%$ on the other splits.}
        \label{fig:modification_breakdown}
    \end{subfigure}

    \caption{\textbf{Evolved-skill usage and workflow-edit behavior of \ours{} on Claude-Sonnet-4.5} (aggregated over LongCat-Image and Z-Image-Turbo).
    \emph{Left:} agents read evolved skills heavily on dense / compositional benchmarks (DPG, GenEval2) and predefined skills more on the OneIG splits.
    \emph{Right:} only $\sim\!39\%$ of edits are prompt-text rewrites; the agent spends the rest on hyperparameters, graph topology, model/LoRA choices, and multi-pass design, showing that \ours{} performs broad workflow construction rather than pure prompt engineering.}
    \label{fig:combined_results}
\end{figure}
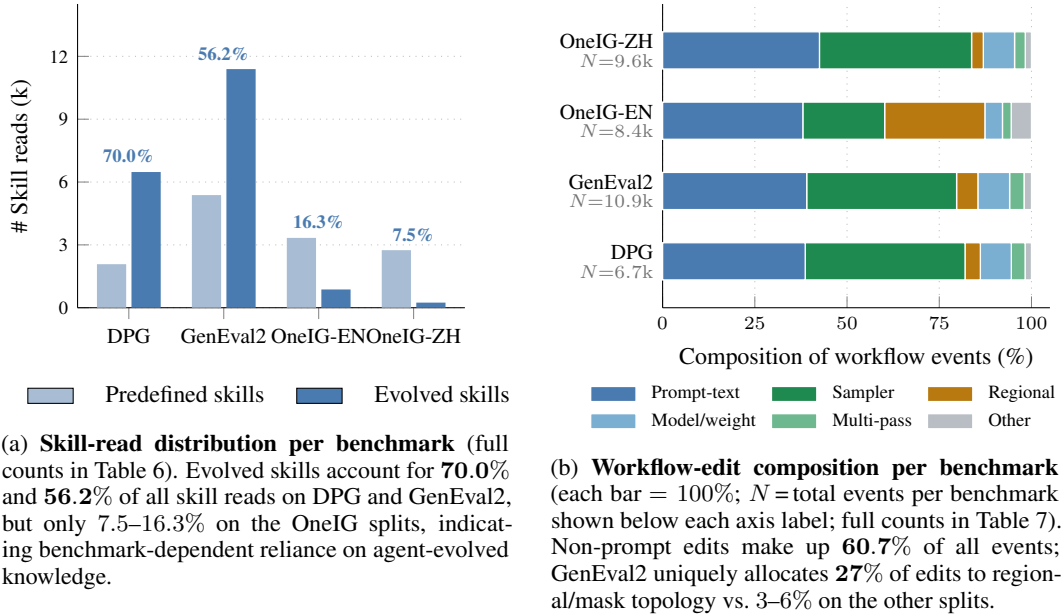

\textbf{Evolved skills capture reusable patterns for recurring image-generation workflows.}
Table~\ref{tab:evolved-skill-examples} shows representative evolved skills from different benchmarks.
Each skill is a named \texttt{SKILL.md} procedure that captures a recurring workflow-execution pattern rather than a one-off memory.
For GenEval2, skills such as \texttt{spatial-anchor-with-count}, \texttt{spatial-count-binding}, and \texttt{attribute-binding} encode reusable procedures for preserving object counts, attribute bindings, and spatial layouts in compositional scenes.
For DPG-Bench, skills such as \texttt{material-texture-detail}, \texttt{precise-color-attribution}, and \texttt{lighting-and-reflection-detail} focus on fine-grained control of material appearance, color assignment, and illumination.
For OneIG-EN and OneIG-ZH, the learned skills shift toward anime-specific character and style control, including \texttt{character-counting}, \texttt{anime-danbooru-ordering}, and \texttt{anime-single-character-simple}.
These examples show that evolved skills are not generic memories, but benchmark and style-specific reusable procedures for recurring workflow problems.
\begin{table}[t]
\centering
\small
\begin{tcolorbox}[
  width=\linewidth,
  colback=skillbg,
  colframe=black,
  boxrule=0.7pt,
  arc=2mm,
  left=6pt,
  right=6pt,
  top=6pt,
  bottom=6pt
]
\centering

\colorbox{skilltitle}{%
  \parbox{\dimexpr\linewidth-2\fboxsep\relax}{\centering\bfseries
  Representative Evolved Skills}
}

\vspace{6pt}

\colorbox{skillsection}{%
  \parbox{\dimexpr\linewidth-2\fboxsep\relax}{\textbf{Representative skills by benchmark}}
}

\vspace{3pt}
\renewcommand{\arraystretch}{1.16}
\setlength{\tabcolsep}{5pt}
\rowcolors{2}{skillalt}{white}
\begin{tabularx}{\linewidth}{@{}p{0.17\linewidth}p{0.35\linewidth}X@{}}
\toprule
\rowcolor{white}
\textbf{Benchmark} & \textbf{Example skills} & \textbf{What they capture} \\
\midrule
GenEval2 &
\shortstack[l]{\texttt{spatial-anchor-with-count}\\
\texttt{spatial-count-binding}\\
\texttt{attribute-binding}}
&
Reusable procedures for preserving object counts, binding attributes to the correct object, and maintaining spatial layouts in compositional scenes. \\

DPG-Bench &
\shortstack[l]{\texttt{material-texture-detail}\\
\texttt{precise-color-attribution}\\
\texttt{lighting-and-reflection-detail}}
&
Reusable procedures for controlling fine-grained material appearance, color assignment, and lighting or reflection cues. \\

OneIG-EN &
\shortstack[l]{\texttt{character-counting}\\
\texttt{anime-danbooru-ordering}\\
\texttt{anime-style-declaration}}
&
Reusable procedures for enforcing anime-style prompt structure and controlling the number and arrangement of characters. \\

OneIG-ZH &
\shortstack[l]{\texttt{anime-single-character-simple}\\
\texttt{anime-multi-character}\\
\texttt{resolution-quality-tags}}
&
Reusable procedures for character-layout control and benchmark-specific style and resolution conventions. \\
\bottomrule
\end{tabularx}
\rowcolors{2}{}{}
\end{tcolorbox}

\caption{Representative evolved skills in \ours{}. The learned skills are benchmark-specific reusable procedures rather than raw memories, capturing recurring workflow-execution patterns such as count and spatial binding, fine-grained appearance control, and anime-specific character prompting.}
\label{tab:evolved-skill-examples}
\end{table}

\textbf{Evolved skills are heavily used by agents.}
We analyze the skills evolved by Claude-Sonnet-4.5 across four benchmarks and two image-generation models.
As shown in Figure~\ref{fig:evolved_reads}, evolved skills account for an average of $\sim\!50\%$ of total skill reads across benchmarks (specifically $70.0\%$ on DPG-Bench and $56.2\%$ on GenEval2, dropping to $16.3\%$ on OneIG-EN and $7.5\%$ on OneIG-ZH), indicating that agents rely heavily on them when constructing image-generation workflows for dense and compositional prompts.
Across the eight experimental settings, Claude-Sonnet-4.5 produced $318$ unique skills, totaling $4{,}768$ skill versions.
These skills reflect the distinct requirements of each benchmark.
GenEval2 primarily induces object-counting and attribute-binding skills; DPG-Bench emphasizes spatial relations, poses, and anatomy; OneIG-EN favors anime-style prompting and character-count control; and OneIG-ZH further introduces Chinese-specific anime conventions and resolution-tag skills.\footnote{The top ten evolved skills are listed in Appendix~\ref{app:evolved_skills}.}
The evolved skills provide reusable, task-specific patterns that help agents start from stronger knowledge and generate higher-quality images.

\textbf{Agents perform more complex actions than simple prompt rewriting.}
We track the actions taken by the Claude agent when constructing workflows in \ours{}, as shown in Figure~\ref{fig:modification_breakdown}.
Across $35{,}612$ workflow events, prompt-text edits are the most frequent single category but account for only $39.3\%$ of all edits.
The remaining $60.7\%$ modify other parts of the workflow, including sampler and guidance hyperparameters ($35.8\%$), regional or mask-based graph topology ($11.4\%$, with a striking $27\%$ on GenEval2), LoRA / checkpoint selection ($7.3\%$), and multi-pass upscaling ($3.2\%$).
This shows that \ours{} improves image generation through broader workflow optimization rather than relying only on prompt refinement.



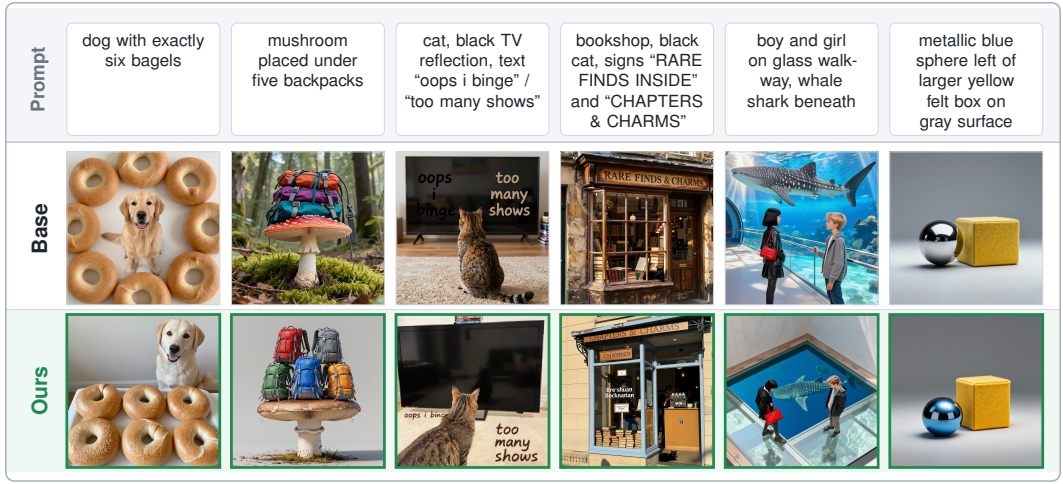
\begin{figure}[!htbp]
\centering
\definecolor{ccOursGreen}{HTML}{1F8A4F}
\definecolor{ccOursTint}{HTML}{EFF8F3}
\definecolor{ccHeader}{HTML}{F3F5F8}
\definecolor{ccCell}{HTML}{FFFFFF}
\definecolor{ccStripe}{HTML}{F7F8FA}
\definecolor{ccRule}{HTML}{C9CCD1}
\definecolor{ccCellBd}{HTML}{D2D7DC}
\definecolor{ccOuterBd}{HTML}{A8B0BA}
\definecolor{ccText}{HTML}{1F252D}
\definecolor{ccTextDim}{HTML}{6B7280}
\definecolor{ccTag}{HTML}{1F8A4F}

\setlength{\fboxsep}{0pt}%
\resizebox{\linewidth}{!}{%
\begin{tikzpicture}[
    font=\sffamily,
    inner sep=0pt, outer sep=0pt,
    every node/.style={inner sep=0pt, outer sep=0pt},
]
\def\imgW{1.58}          
\def\imgH{1.58}          
\def\xA{0.55}
\def\xB{2.25}
\def\xC{3.95}
\def\xD{5.65}
\def\xE{7.35}
\def\xF{9.05}
\def\xR{10.73}           
\def\yTop{0.08}
\def\yHB{-1.30}          
\def\yBase{-1.40}        
\def\yOurs{-3.08}        
\def\yBot{-4.76}         

\fill[white]     (-0.08,\yTop) rectangle (\xR+0.08,\yBot-0.06);
\fill[ccHeader]  (-0.08,\yTop) rectangle (\xR+0.08,\yHB);
\fill[ccOursTint](-0.08,\yOurs+0.05) rectangle (\xR+0.08,\yOurs-\imgH-0.05);

\draw[ccRule, line width=0.50pt] (-0.08,\yHB) -- (\xR+0.08,\yHB);
\draw[ccCellBd, line width=0.28pt] (-0.08,\yBase-\imgH-0.05) -- (\xR+0.08,\yBase-\imgH-0.05);

\def\casehead#1#2#3{%
  \fill[ccCell, rounded corners=1.5pt]
    (#1,-0.07) rectangle ++(\imgW,-1.16);%
  \draw[ccCellBd, line width=0.22pt, rounded corners=1.5pt]
    (#1,-0.07) rectangle ++(\imgW,-1.16);%
  \node[anchor=north, text width=1.46cm, align=center,
        font=\sffamily\fontsize{4.9pt}{6.0pt}\selectfont, text=ccText]
    at (#1+0.5*\imgW,-0.18) {#3};%
}

\def\methodlabel#1#2#3{%
  \node[font=\sffamily\bfseries\fontsize{6.2pt}{7.2pt}\selectfont, text=#3]
    at (0.27,#1-0.5*\imgH) {\rotatebox{90}{#2}};%
}

\def\imgcell#1#2#3{%
  \node[anchor=north west] at (#1,#2)
    {\includegraphics[width=\imgW cm]{figures/comfyclaw_cherry_pick/thumbs/#3.jpg}};%
  \draw[ccCellBd, line width=0.25pt] (#1,#2) rectangle ++(\imgW,-\imgH);%
}

\def\ourscell#1#2#3{%
  \node[anchor=north west] at (#1,#2)
    {\includegraphics[width=\imgW cm]{figures/comfyclaw_cherry_pick/thumbs/#3.png}};%
  \draw[ccOursGreen, line width=0.80pt] (#1,#2) rectangle ++(\imgW,-\imgH);%
}

\node[font=\sffamily\bfseries\fontsize{5.5pt}{6.5pt}\selectfont, text=ccTextDim, rotate=90]
  at (0.27,-0.65) {Prompt};

\casehead{\xA}{COUNTING}{dog with exactly six bagels}
\casehead{\xB}{COMPOSITION}{mushroom placed under five backpacks}
\casehead{\xC}{TEXT \& MIRROR}{cat, black TV reflection, text ``oops i binge'' / ``too many shows''}
\casehead{\xD}{SCENE TEXT}{bookshop, black cat, signs ``RARE FINDS INSIDE'' and ``CHAPTERS \& CHARMS''}
\casehead{\xE}{STYLIZATION}{boy and girl on glass walkway, whale shark beneath}
\casehead{\xF}{SPATIAL}{metallic blue sphere left of larger yellow felt box on gray surface}

\methodlabel{\yBase}{Base}{ccText}
\methodlabel{\yOurs}{Ours}{ccOursGreen}

\imgcell{\xA}{\yBase}{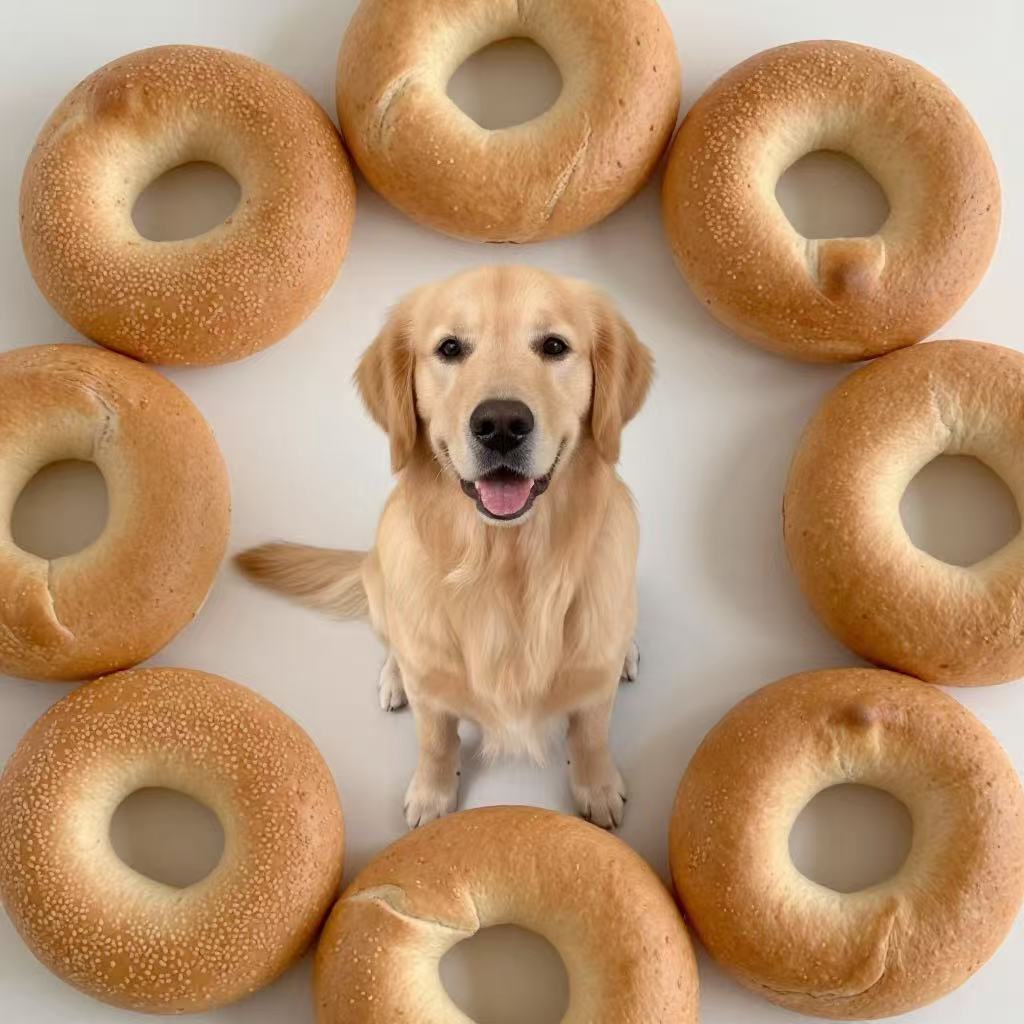}
\imgcell{\xB}{\yBase}{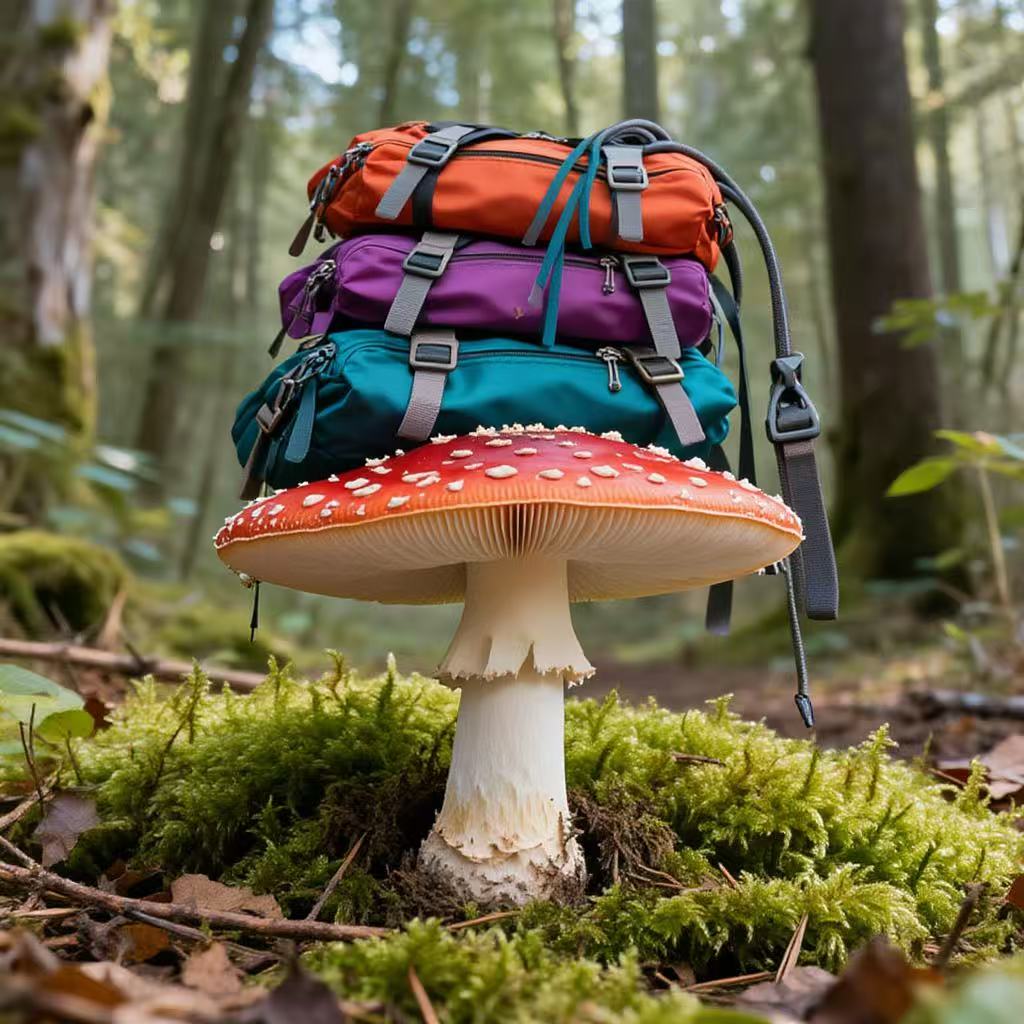}
\imgcell{\xC}{\yBase}{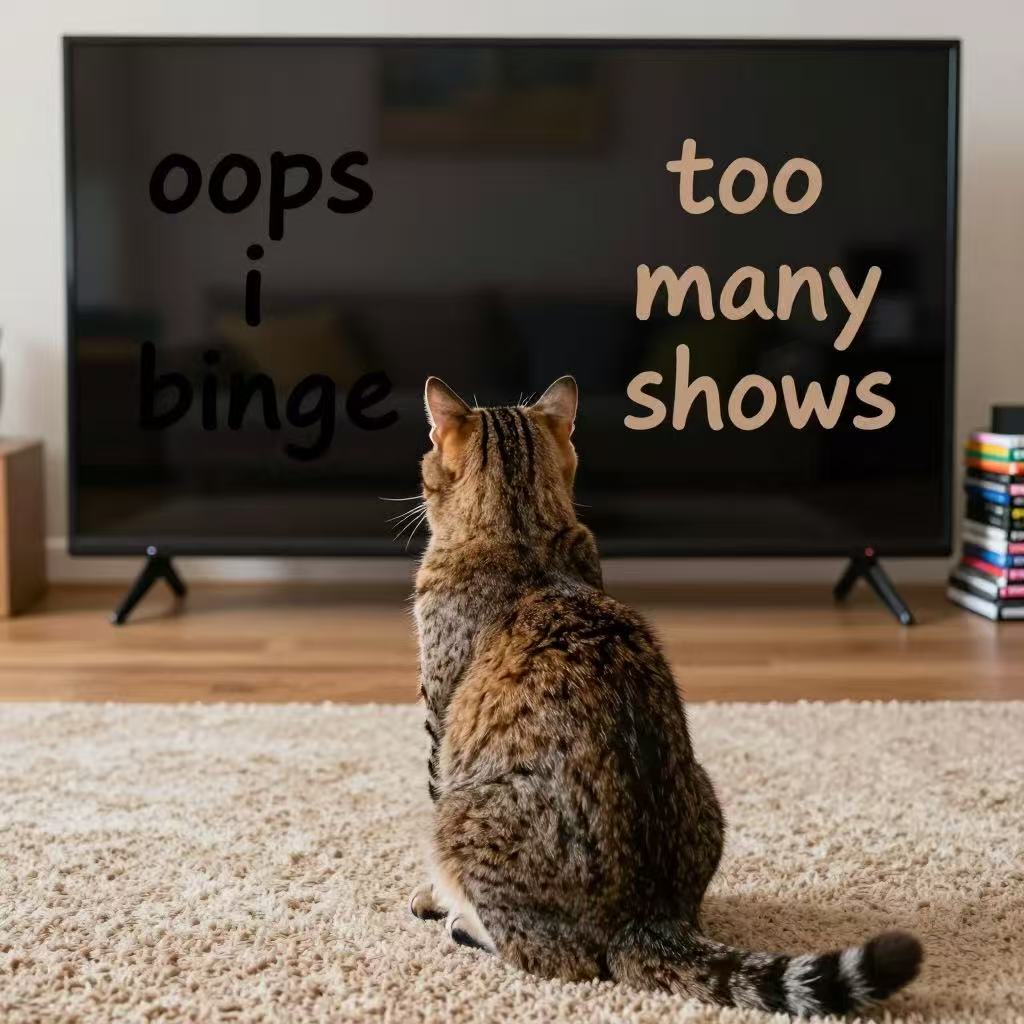}
\imgcell{\xD}{\yBase}{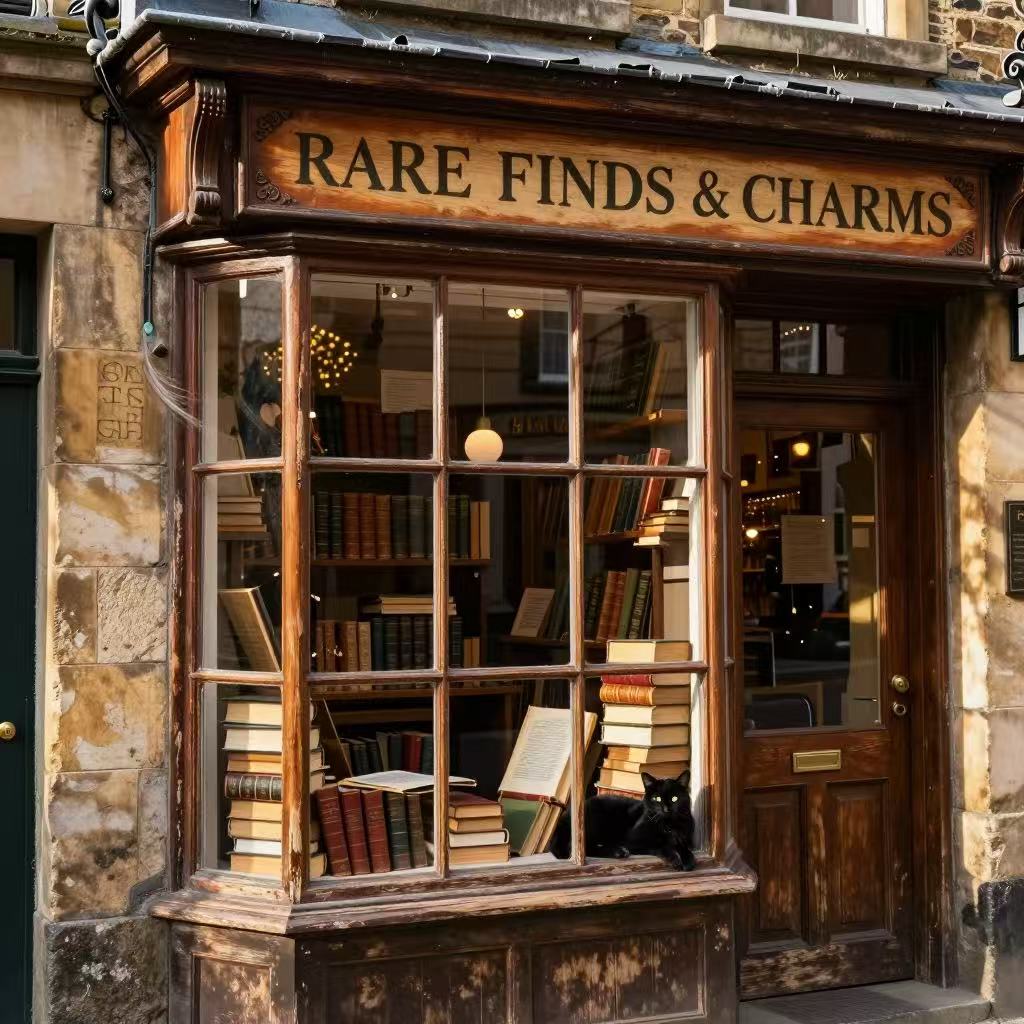}
\imgcell{\xE}{\yBase}{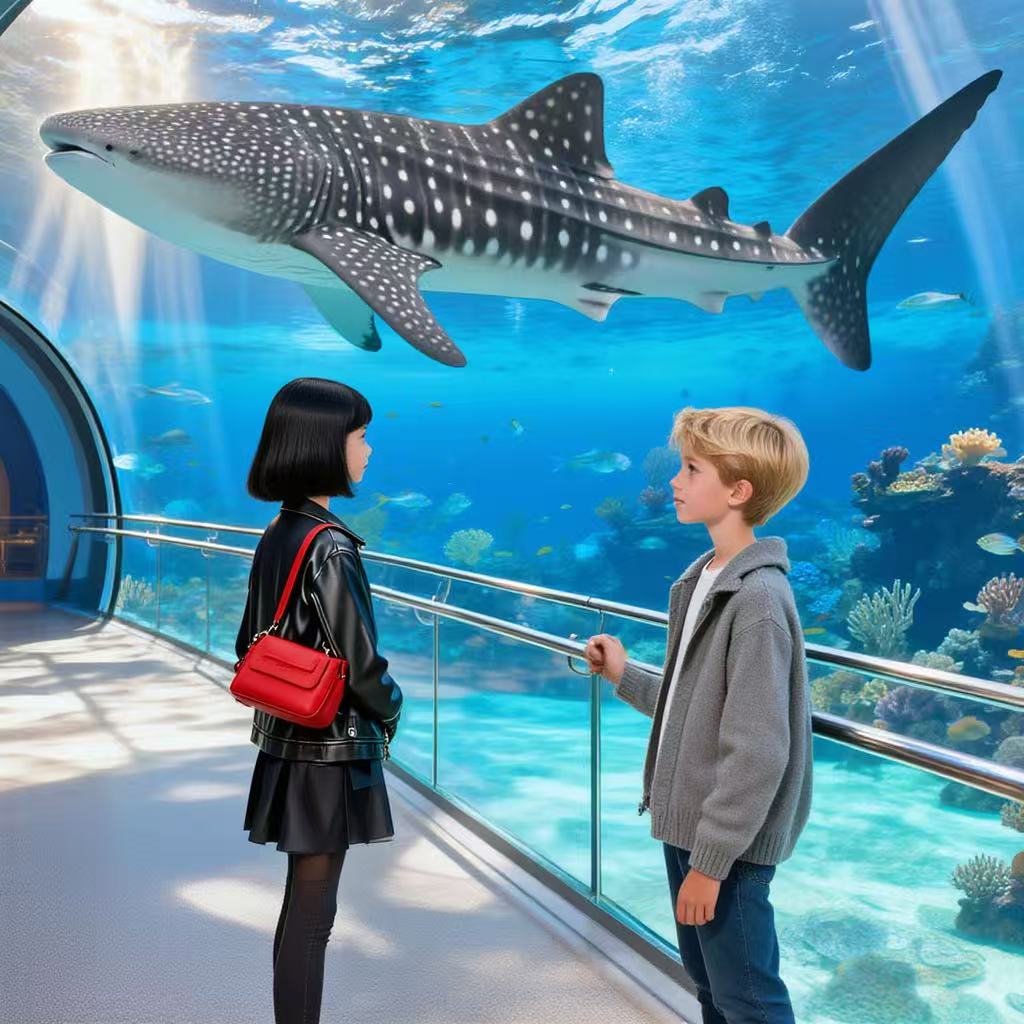}
\imgcell{\xF}{\yBase}{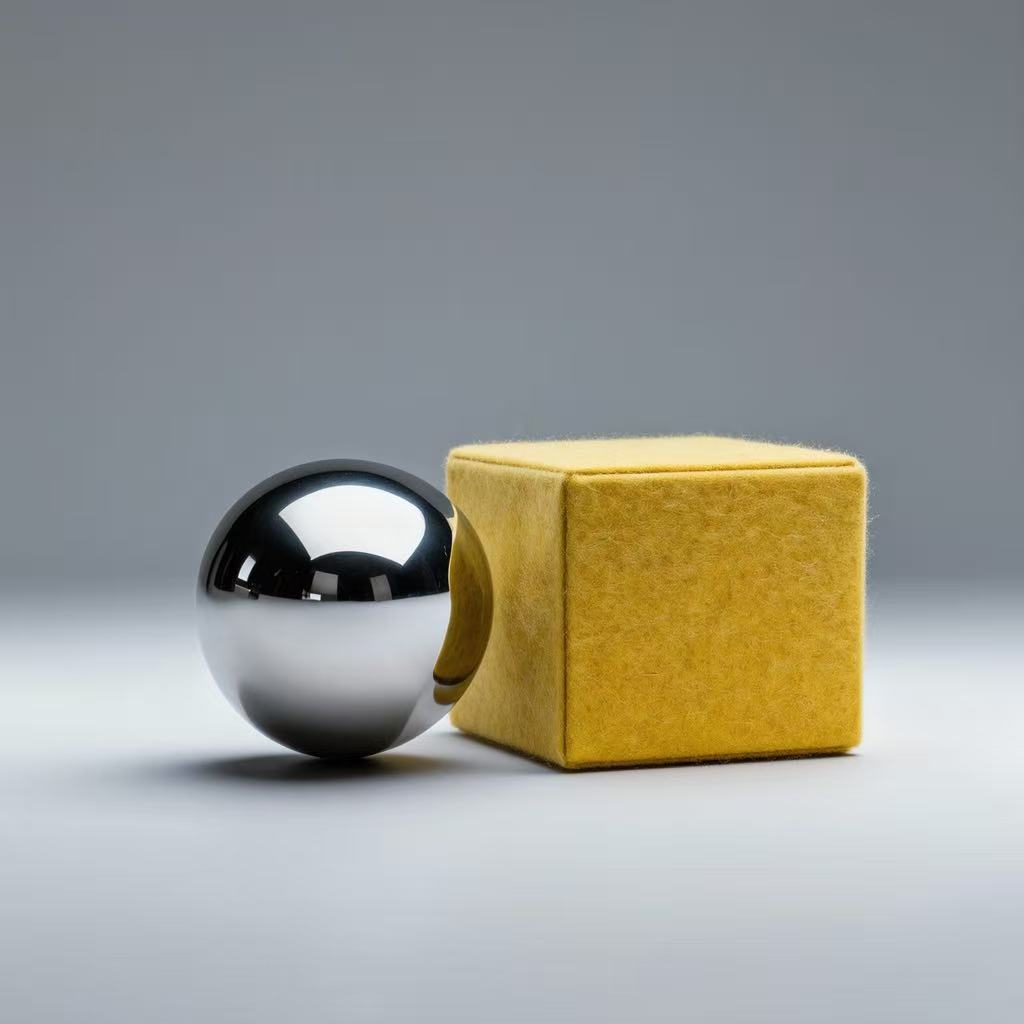}

\ourscell{\xA}{\yOurs}{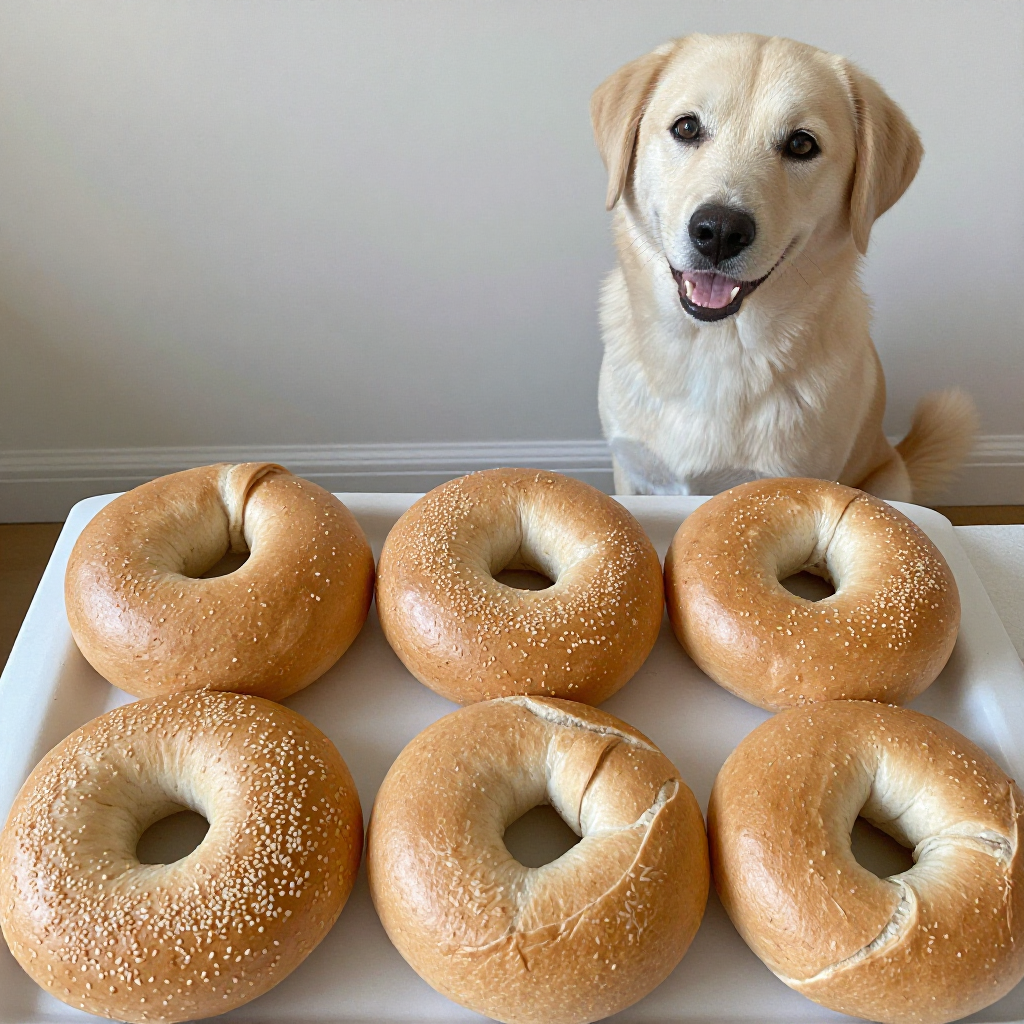}
\ourscell{\xB}{\yOurs}{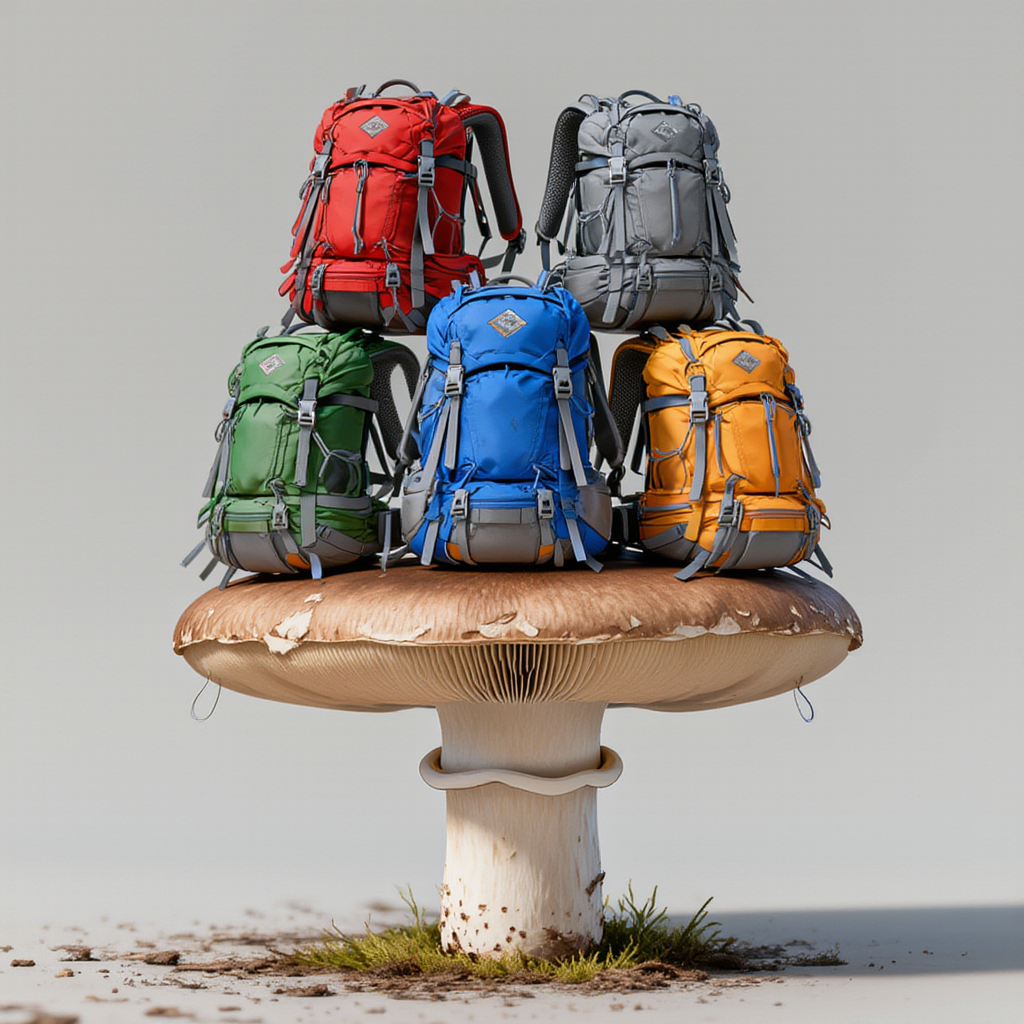}
\ourscell{\xC}{\yOurs}{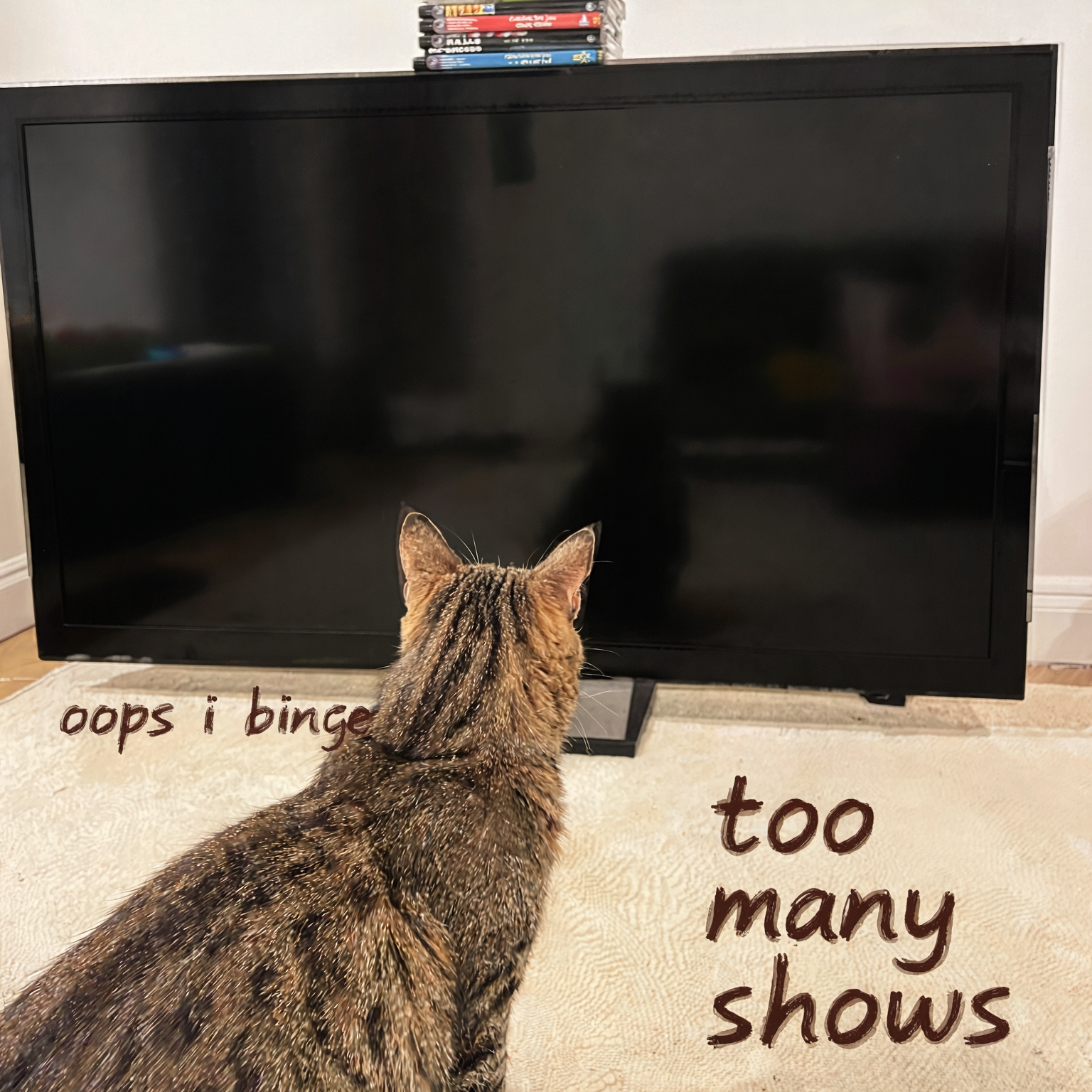}
\ourscell{\xD}{\yOurs}{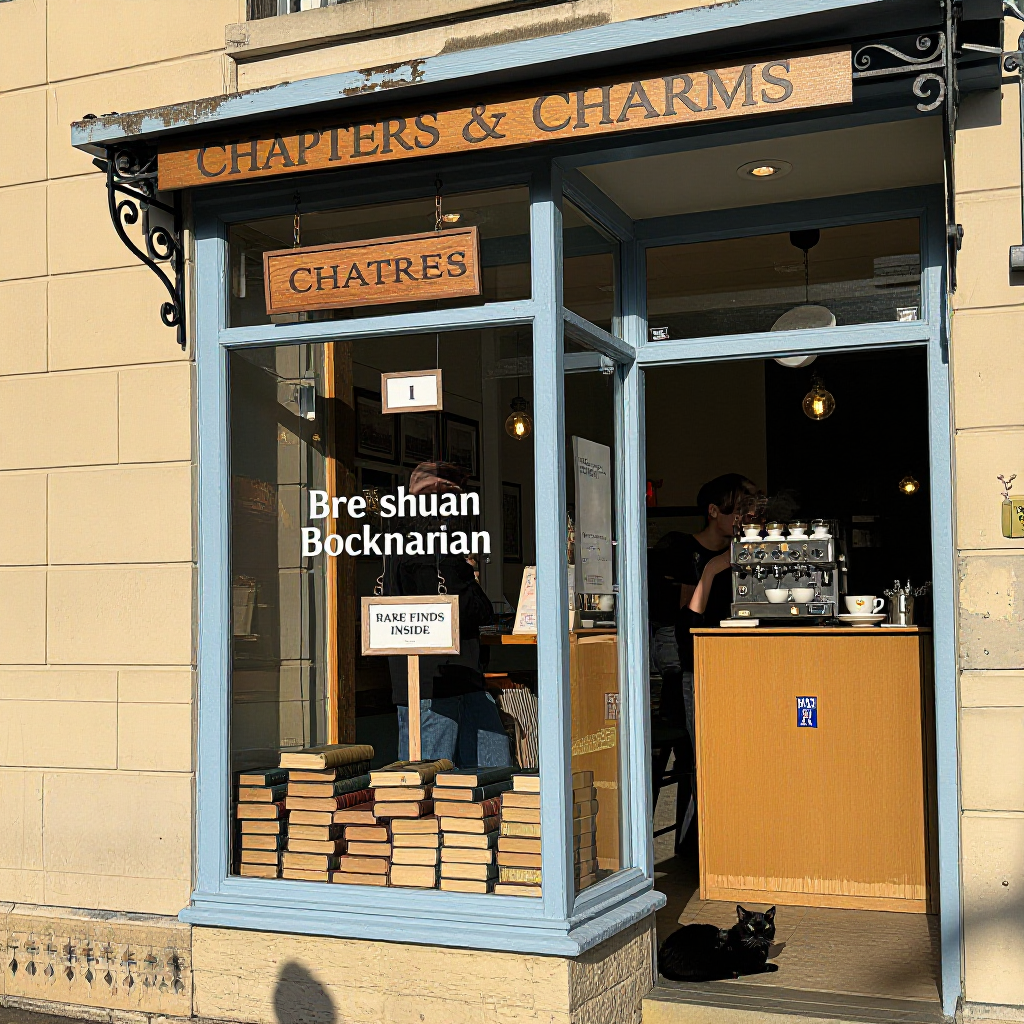}
\ourscell{\xE}{\yOurs}{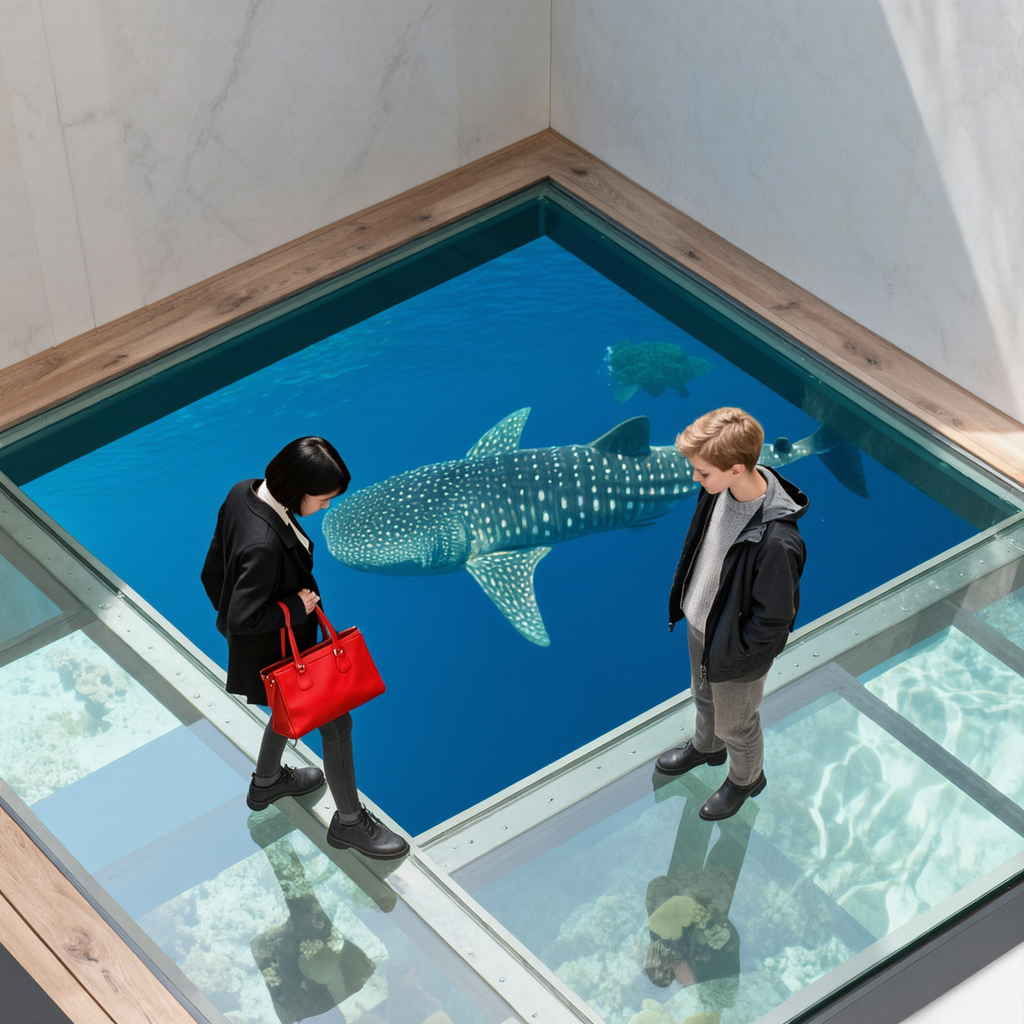}
\ourscell{\xF}{\yOurs}{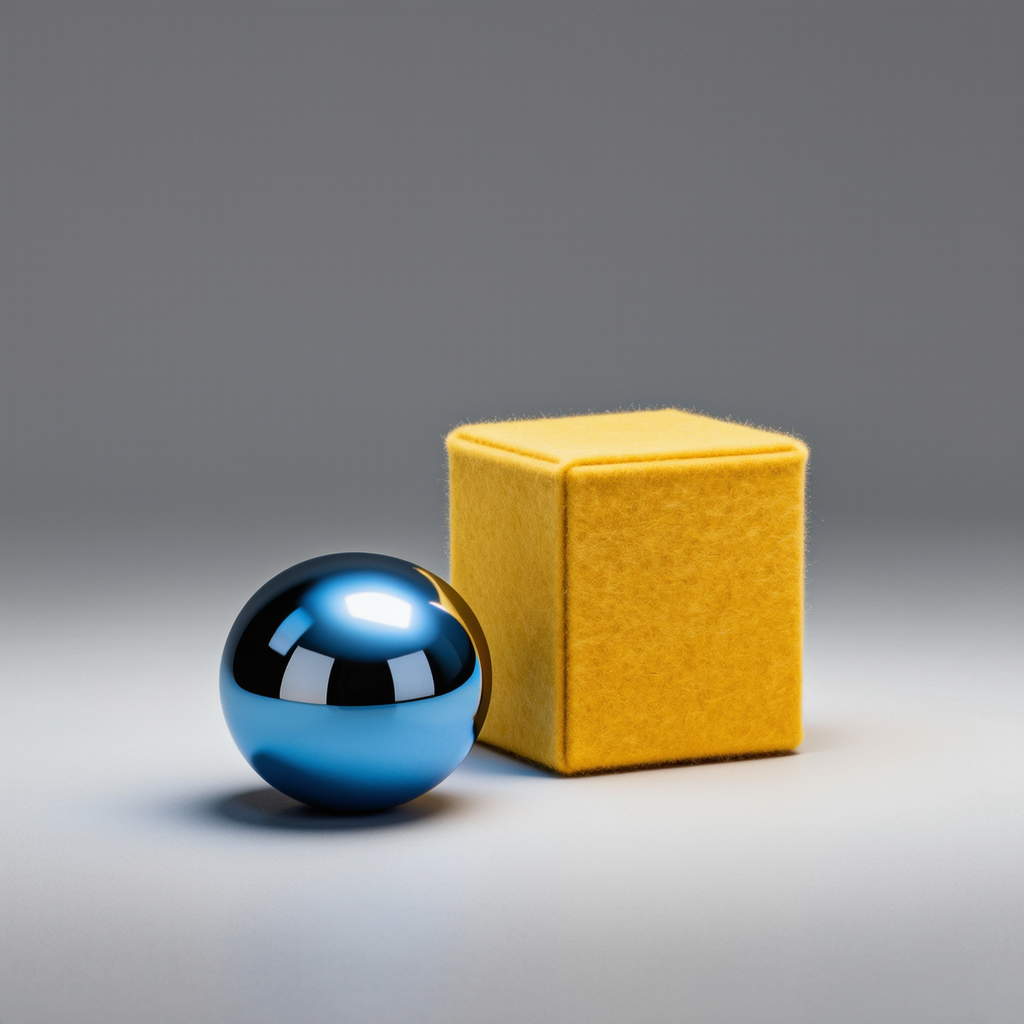}

\draw[ccOuterBd, rounded corners=2.5pt, line width=0.45pt]
  (-0.08,\yTop+0.06) rectangle (\xR+0.08,\yBot-0.04);

\end{tikzpicture}%
}
\vspace{-6pt}
\caption{\textbf{Qualitative comparison across methods on six prompts spanning five capability categories.}
Each column is a prompt (header shows category and full description);
rows are \emph{Base} (single-pass baseline) and \emph{Ours} (\ours{}, green border).
\ours{} more reliably realises object counts, spatial relations, scene-text accuracy, and fine-grained attribute control.}
\label{fig:comfyclaw-cherrypick}
\end{figure}

\FloatBarrier

\subsection{Qualitative Analysis}
We further conduct a qualitative analysis of the generated images, since quantitative metrics alone may not fully capture visual quality in image generation tasks.
Our goal is to compare how different harness designs affect the agent's ability to control the workflow and produce high-quality images.
We hired two annotators to rate image quality on a 1--5 Likert scale, where higher scores indicate better visual quality and stronger text-image alignment.
For each benchmark and each agent, we randomly sample 50 finalized images.
Annotators were shown the input prompt together with the generated images from different experimental groups and asked to assign a Likert rating to each image.\footnote{Annotation cost and instruction details are in Appendix~\ref{app:user_study}.}
In total, 2{,}400 images are annotated.
We show the results in Table~\ref{tab:qualitative_evals}.
Overall, images generated using \ours{} align more with input prompts and exhibit greater visual realism and aesthetics.

\begin{table}[!htbp]
  \centering
  \small
  \resizebox{0.8\columnwidth}{!}{%
  \begin{tabular}{llccccc}
    \toprule
    Image Model & Method & GenEval2  & DPG-Bench  & OneIG-EN  & OneIG-ZH  & Avg \\
    \midrule
    \multirow{3}{*}{\texttt{longcat}} & \baseline{} & 3.63 & 3.81 & 3.50 & 3.64 & 3.65 \\
    & \ours{} & \textbf{4.02} & \textbf{4.74} & \textbf{4.55} & \textbf{4.66} & \textbf{4.49} \\
    \midrule
    \multirow{3}{*}{\texttt{z-image-turbo}} & \baseline{} & 3.58 & 4.54 & 4.42 & 4.48 & 4.26 \\
     & \ours{} & \textbf{4.36} & 4.60 & \textbf{4.59} & \textbf{4.55} & \textbf{4.53} \\
    \bottomrule
  \end{tabular}
  }
  \caption{\textbf{Qualitative evaluation of Claude as an agent to control the ComfyUI workflow.} Overall \ours{} has higher average human visual annotation scores than \comfygems{} and \baseline{}.}
  \label{tab:qualitative_evals}
  \vspace{-10pt}
\end{table}

\begin{figure}[!htbp]
\centering
\definecolor{refOrangeBd}{HTML}{D69220}
\definecolor{refOrangeBg}{HTML}{FCF1DD}
\definecolor{refBlueBd}{HTML}{4A7BB0}
\definecolor{refBlueBg}{HTML}{EAF1FB}
\definecolor{refGreenBd}{HTML}{1F8A4F}
\definecolor{refGreenBg}{HTML}{E9F5EE}
\definecolor{refGreenDot}{HTML}{167040}
\definecolor{refRule}{HTML}{C9CCD1}
\definecolor{refLabel}{HTML}{2A2F36}
\definecolor{refMuted}{HTML}{6B7280}
\setlength{\fboxsep}{0pt}%
\resizebox{\linewidth}{!}{%
\begin{tikzpicture}[
    font=\sffamily,
    inner sep=0pt, outer sep=0pt,
    every node/.style={inner sep=0pt, outer sep=0pt},
    instr/.style={
        draw=refOrangeBd, fill=refOrangeBg, rounded corners=2pt,
        line width=0.4pt,
        text width=1.40cm, align=left,
        minimum height=0.62cm,
        font=\sffamily\tiny\itshape, text=refLabel,
        inner xsep=2pt, inner ysep=2pt,
    },
    crit/.style={
        draw=refBlueBd, fill=refBlueBg, rounded corners=2pt,
        line width=0.4pt,
        text width=1.40cm, align=left,
        minimum height=0.62cm,
        font=\sffamily\tiny, text=refLabel,
        inner xsep=2pt, inner ysep=2pt,
    },
    critGood/.style={
        draw=refGreenBd, fill=refGreenBg, rounded corners=2pt,
        line width=0.4pt,
        text width=1.40cm, align=left,
        minimum height=0.62cm,
        font=\sffamily\tiny, text=refLabel,
        inner xsep=2pt, inner ysep=2pt,
    },
    arrow/.style={->, line width=0.5pt, draw=refMuted},
    arrowlbl/.style={font=\sffamily\tiny, text=refMuted, fill=white, inner sep=0.6pt},
    panel/.style={inner sep=0pt, outer sep=0pt},
    badge/.style={
        draw=refRule, fill=white, rounded corners=1.2pt, line width=0.3pt,
        font=\sffamily\bfseries\tiny, text=refLabel,
        inner xsep=2pt, inner ysep=1pt,
    },
    badgeFinal/.style={
        draw=refGreenBd, fill=refGreenBg, rounded corners=1.2pt, line width=0.4pt,
        font=\sffamily\bfseries\tiny, text=refGreenDot,
        inner xsep=2pt, inner ysep=1pt,
    },
    rowtitle/.style={
        anchor=north west,
        font=\sffamily\bfseries\footnotesize, text=refLabel,
    },
]
\def\imgW{1.40}        
\def\imgH{1.40}        
\def\inGap{0.25}       
\def\midGap{0.65}      

\def\xSa{0.00}
\def\xAa{1.65}         
\def\xBa{3.30}
\def\xCa{4.95}
\def\xRa{6.35}         

\def\xSb{7.00}         
\def\xAb{8.65}
\def\xBb{10.30}
\def\xCb{11.95}
\def\xRb{13.35}        

\def\xSCa{0.70} \def\xACa{2.35} \def\xBCa{4.00} \def\xCCa{5.65}
\def\xSCb{7.70} \def\xACb{9.35} \def\xBCb{11.00} \def\xCCb{12.65}

\def\xArAa{1.525} \def\xArBa{3.175} \def\xArCa{4.825}
\def\xArAb{8.525} \def\xArBb{10.175} \def\xArCb{11.825}

\def\xMid{6.675}       

\def\yTitle{-0.05}     
\def\yOrgTop{-0.45}    
\def\yImgTop{-1.20}    
\def\yImgMid{-1.90}    
\def\yImgBot{-2.60}    
\def\yBluTop{-2.70}    
\def\yBot{-3.40}       

\fill[white] (-0.20, 0.25) rectangle (\xRb + 0.20, \yBot);

\draw[densely dashed, refRule, line width=0.4pt]
    (\xMid, +0.10) -- (\xMid, \yBot + 0.05);

\node[rowtitle] at (\xSa, \yTitle)
    {(a)~\textit{four purple lions}.};

\node[instr, anchor=north] at (\xArAa, \yOrgTop)
    {Force unusual color.};
\node[instr, anchor=north] at (\xArBa, \yOrgTop)
    {Add regional control.};
\node[instr, anchor=north] at (\xArCa, \yOrgTop)
    {Stack Z-Image LoRAs.};

\node[panel, anchor=north west] at (\xSa, \yImgTop)
    {\includegraphics[width=\imgW cm, height=\imgH cm]{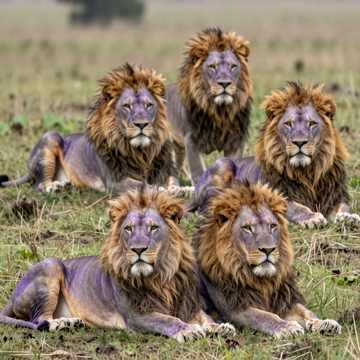}};
\node[panel, anchor=north west] at (\xAa, \yImgTop)
    {\includegraphics[width=\imgW cm, height=\imgH cm]{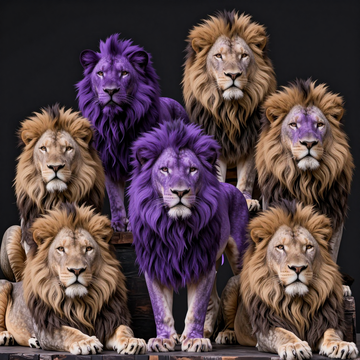}};
\node[panel, anchor=north west] at (\xBa, \yImgTop)
    {\includegraphics[width=\imgW cm, height=\imgH cm]{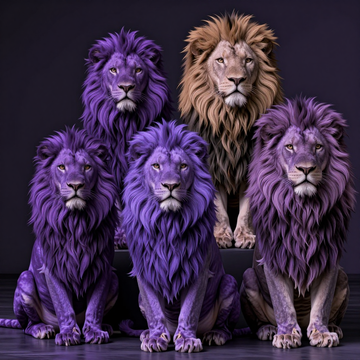}};
\node[panel, anchor=north west] at (\xCa, \yImgTop)
    {\includegraphics[width=\imgW cm, height=\imgH cm]{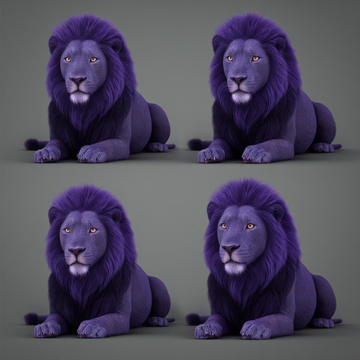}};

\draw[refRule, line width=0.4pt] (\xSa, \yImgTop) rectangle ++(\imgW, -\imgH);
\draw[refRule, line width=0.4pt] (\xAa, \yImgTop) rectangle ++(\imgW, -\imgH);
\draw[refRule, line width=0.4pt] (\xBa, \yImgTop) rectangle ++(\imgW, -\imgH);
\draw[refGreenBd, line width=0.85pt] (\xCa, \yImgTop) rectangle ++(\imgW, -\imgH);

\node[badge, anchor=north west]      at (\xSa+0.04, \yImgTop-0.04) {Initial};
\node[badge, anchor=north west]      at (\xAa+0.04, \yImgTop-0.04) {Iter 1};
\node[badge, anchor=north west]      at (\xBa+0.04, \yImgTop-0.04) {Iter 2};
\node[badgeFinal, anchor=north west] at (\xCa+0.04, \yImgTop-0.04) {Final};

\draw[arrow] (\xSa+\imgW+0.04, \yImgMid) -- (\xAa-0.04, \yImgMid);
\draw[arrow] (\xAa+\imgW+0.04, \yImgMid) -- (\xBa-0.04, \yImgMid);
\draw[arrow] (\xBa+\imgW+0.04, \yImgMid) -- (\xCa-0.04, \yImgMid);
\node[arrowlbl] at (\xArAa, \yImgMid+0.13) {};
\node[arrowlbl] at (\xArBa, \yImgMid+0.13) {};
\node[arrowlbl] at (\xArCa, \yImgMid+0.13) {};

\node[crit, anchor=north]     at (\xSCa, \yBluTop)
    {Lions tan, not purple.};
\node[crit, anchor=north]     at (\xACa, \yBluTop)
    {6 lions; 2 purple.};
\node[crit, anchor=north]     at (\xBCa, \yBluTop)
    {5 lions; 1 still tan.};
\node[critGood, anchor=north] at (\xCCa, \yBluTop)
    {4 lions, all purple.};

\node[rowtitle] at (\xSb, \yTitle)
    {(b)~\textit{clock left of three glass pigs}.};

\node[instr, anchor=north] at (\xArAb, \yOrgTop)
    {Split regions + LoRA.};
\node[instr, anchor=north] at (\xArBb, \yOrgTop)
    {Drop LoRA; force glass.};
\node[instr, anchor=north] at (\xArCb, \yOrgTop)
    {Simplify regions; count.};

\node[panel, anchor=north west] at (\xSb, \yImgTop)
    {\includegraphics[width=\imgW cm, height=\imgH cm]{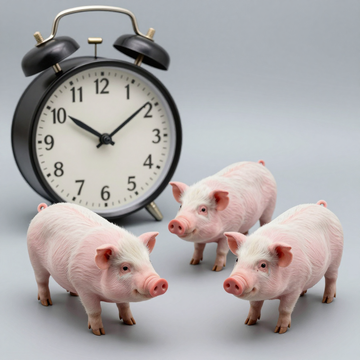}};
\node[panel, anchor=north west] at (\xAb, \yImgTop)
    {\includegraphics[width=\imgW cm, height=\imgH cm]{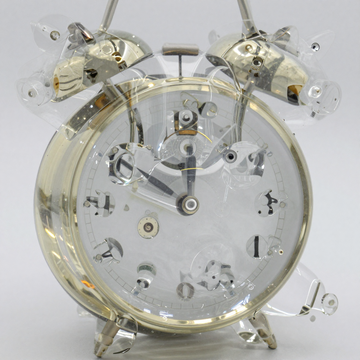}};
\node[panel, anchor=north west] at (\xBb, \yImgTop)
    {\includegraphics[width=\imgW cm, height=\imgH cm]{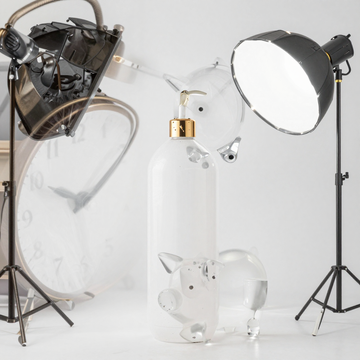}};
\node[panel, anchor=north west] at (\xCb, \yImgTop)
    {\includegraphics[width=\imgW cm, height=\imgH cm]{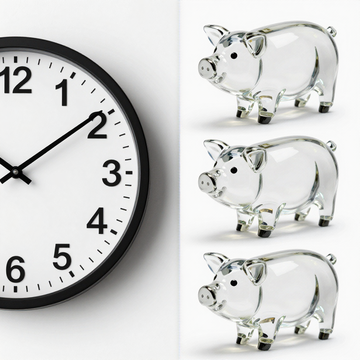}};

\draw[refRule, line width=0.4pt] (\xSb, \yImgTop) rectangle ++(\imgW, -\imgH);
\draw[refRule, line width=0.4pt] (\xAb, \yImgTop) rectangle ++(\imgW, -\imgH);
\draw[refRule, line width=0.4pt] (\xBb, \yImgTop) rectangle ++(\imgW, -\imgH);
\draw[refGreenBd, line width=0.85pt] (\xCb, \yImgTop) rectangle ++(\imgW, -\imgH);

\node[badge, anchor=north west]      at (\xSb+0.04, \yImgTop-0.04) {Initial};
\node[badge, anchor=north west]      at (\xAb+0.04, \yImgTop-0.04) {Iter 1};
\node[badge, anchor=north west]      at (\xBb+0.04, \yImgTop-0.04) {Iter 2};
\node[badgeFinal, anchor=north west] at (\xCb+0.04, \yImgTop-0.04) {Final};

\draw[arrow] (\xSb+\imgW+0.04, \yImgMid) -- (\xAb-0.04, \yImgMid);
\draw[arrow] (\xAb+\imgW+0.04, \yImgMid) -- (\xBb-0.04, \yImgMid);
\draw[arrow] (\xBb+\imgW+0.04, \yImgMid) -- (\xCb-0.04, \yImgMid);
\node[arrowlbl] at (\xArAb, \yImgMid+0.13) {};
\node[arrowlbl] at (\xArBb, \yImgMid+0.13) {};
\node[arrowlbl] at (\xArCb, \yImgMid+0.13) {};

\node[crit, anchor=north]     at (\xSCb, \yBluTop)
    {Pigs not glass.};
\node[crit, anchor=north]     at (\xACb, \yBluTop)
    {Pigs gone; clock only.};
\node[crit, anchor=north]     at (\xBCb, \yBluTop)
    {Glass OK; clock lost.};
\node[critGood, anchor=north] at (\xCCb, \yBluTop)
    {Clock left, 3 glass pigs.};

\draw[refRule, rounded corners=2pt, line width=0.4pt]
    (-0.20, 0.25) rectangle (\xRb + 0.20, \yBot);

\end{tikzpicture}%
}
\caption{\textbf{Iterative visual refinement under \ours{}.} (a) \emph{four purple lions} (verifier $0.33{\to}0.91$); (b) \emph{a clock to the left of three glass pigs} (verifier $0.31{\to}0.96$). Each strip reads left to right: the agent inspects the current image, the verifier emits a critique (\textcolor{refBlueBd}{\textbf{blue}}), the agent turns it into a refinement instruction (\textcolor{refOrangeBd}{\textbf{orange}}) for the next pass, and re-executes the workflow; the selected best output is highlighted in \textcolor{refGreenBd}{\textbf{green}}. Refinement repairs unusual-color attributes, conflicting workflow components, or spatial-relation failures that the single-pass baseline could not resolve.}
\label{fig:comfyclaw-refinement}
\vspace{-10pt}
\end{figure}
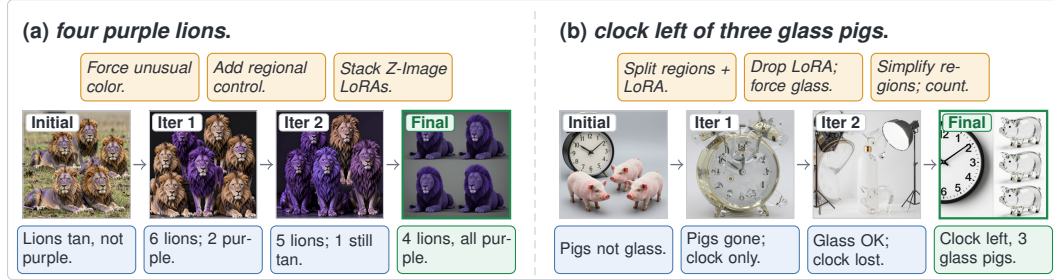

\textbf{Closed-loop workflow refinement repairs failures that
prompt-only rewriting often cannot.}
Figure~\ref{fig:comfyclaw-refinement} shows \ours{}'s refinement loop for editing the workflow.
\textit{First}, the edits are structural rather than purely textual prompt revision.
In Figure~\ref{fig:comfyclaw-refinement}(a), the agent stacks two
Z-Image LoRAs and adds a regional-attention block before the unusual attribute
\emph{purple} is correctly bound to all four lions.
In Figure~\ref{fig:comfyclaw-refinement}(b), the agent adjusts the regional
split until the spatial relation \emph{left of}, the material \emph{glass},
and the count \emph{three} are all satisfied.

\textit{Second}, refinement is not always monotonic.
Some intermediate attempts are worse than the baseline, such as when the pigs
disappear or the clock is lost, but the best-so-far buffer preserves the best
valid candidate while still allowing the agent to learn from failed attempts
and recover in later iterations.

\textit{Third}, the loop is verifier-driven.
Each refinement instruction directly responds to the preceding critique, such
as correcting the lion color or removing an ineffective LoRA after objects
disappear.
Thus, the agent is not retrying blindly; it spends additional graph edits and
re-renders on the specific failures localized by the verifier.
These extra operations improve compositional errors, including
attribute binding, object count, spatial relations, LoRA weight stacking, and hyperparameter adjustments that are difficult for
single-pass prompt-only generation to fix reliably.

\FloatBarrier


\section{Conclusion}
\label{sec:conclusion}
Agentic workflow control is a promising direction for helping users, especially beginners, operate complex tools and automate multi-step workflows.
In image generation, workflow-based systems provide more controllability than prompt-only interfaces by exposing components such as model selection,
conditioning, sampling, refinement, and LoRA adapters.
We propose \ours{}, a self-evolving workflow-control framework that converts past successes and failures into reusable skills, enabling agents to improve how they construct and refine image-generation workflows over time.
Our experiments show that this self-evolving skill mechanism produces higher quality and more realistic images than agent workflow control without skill evolution, and is more preferred by human annotators.
These results suggest that effective visual generation agents should not only execute workflows, but also learn and evolve reusable workflow procedures from their own experience.

\section{Limitations and Future Work}
\label{sec:limitation}

Our current study focuses only on image-generation workflows.
However, ComfyUI also supports video-generation workflows, which require more
complex node control, scene planning, frame consistency, and temporal
conditioning.
Video generation also requires substantially more compute and longer execution
time, making agent control more difficult because long-running workflows can
increase latency, timeout failures, and response errors.
We therefore leave video workflow control as an important direction for future
work.
Future systems will need stronger workflow-control strategies, better skill
management, and more efficient skill retrieval, since video workflows are
typically longer and more complex than image workflows.
Managing the agent context window will also become more challenging, making
compact and accurate skill retrieval especially important.

\bibliography{neurips2026_conference}
\bibliographystyle{unsrtnat}

\newpage
\appendix
\section{Predefined Tools, Skills, and LoRA Settings}
\label{app:tools-skills}

\paragraph{Workflow tools.}
We expose 17 basic tools for controlling ComfyUI workflows.
These tools allow the agent to inspect the current workflow, add and remove
nodes, connect nodes, edit node inputs, set prompts, configure model-specific
parameters, validate the workflow, and submit the final graph for rendering.
Examples include \textsc{add\_node}, \textsc{set\_prompt}, and
\textsc{set\_lora}.
Together, these tools allow the agent to construct and repair workflows at the
graph level rather than only rewriting the text prompt.

\paragraph{Predefined skills.}
In addition to evolved skills, we provide four predefined skills inspired
by~\citet{he2026gems}: \textsc{photorealistic}, \textsc{creative},
\textsc{high-quality}, and \textsc{prompt-artist}.
These skills provide general image-generation guidance, such as improving
photorealism, increasing visual creativity, enhancing overall quality, and
rewriting prompts for stronger visual composition.
They serve as the initial reusable skill library before benchmark-specific
skills are evolved.

\paragraph{Model-specific LoRA settings.}
For \textsc{Z-Image-Turbo}, we include two LoRAs.
The first is a realistic snapshot LoRA~\cite{cai2025z,realisticsnapshot2026},
which encourages more photorealistic and real-life imagery.
The second is an enhancer LoRA~\cite{enhancer2026}, which improves generation
quality across a range of visual styles.
\textsc{LongCat-Image} does not currently support LoRAs in our setup, so
LoRA-based tools are disabled for that model.

\begin{table}[!htbp]
\centering
\small
\setlength{\tabcolsep}{2pt}
\begin{tabular}{@{}p{0.21\linewidth}p{0.38\linewidth}p{0.35\linewidth}@{}}
\toprule
\textbf{Group} & \textbf{Tool} & \textbf{Purpose} \\
\midrule
\multirow{3}{*}{\textit{Inspection}}
  & \texttt{inspect\_workflow}        & Summarise nodes, IDs, classes, inputs of the current graph. \\
  & \texttt{query\_available\_\newline{}models} & List installed checkpoints / LoRAs / UNETs / VAEs / upscalers. \\
  & \texttt{explore\_nodes}           & Query \texttt{/object\_info} and classify nodes by pipeline stage. \\
\midrule
\multirow{4}{*}{\textit{Graph edits}}
  & \texttt{add\_node}      & Append a node; return its new ID. \\
  & \texttt{connect\_nodes} & Wire \emph{src}.slot $\to$ \emph{dst}.input. \\
  & \texttt{delete\_node}   & Remove a node and its incident links. \\
  & \texttt{set\_param}     & Set a scalar input on a specific node. \\
\midrule
\multirow{5}{*}{\textit{Composite macros}}
  & \texttt{set\_prompt}              & Set positive/negative text on every encoder feeding a sampler. \\
  & \texttt{add\_lora\_loader}        & Insert a LoRA between model source and downstream consumers. \\
  & \texttt{add\_regional\_\newline{}attention} & Split conditioning into foreground/background regional prompts. \\
  & \texttt{add\_hires\_fix}          & Add latent upscale + second \texttt{KSampler} + \texttt{VAEDecode}. \\
  & \texttt{add\_inpaint\_pass}       & Add a targeted inpaint pass for a specific region. \\
\midrule
\textit{Skills}
  & \texttt{read\_skill}              & Load a \texttt{SKILL.md} body by name (progressive disclosure). \\
\midrule
\multirow{4}{*}{\textit{Control / meta}}
  & \texttt{report\_evolution\_\newline{}strategy} & Declare the iteration plan and top issue before edits. \\
  & \texttt{validate\_workflow}          & Check the graph for dangling refs, wrong slots, missing outputs. \\
  & \texttt{finalize\_workflow}          & Signal completion (auto-validates; blocks if errors remain). \\
  & \texttt{transition\_stage}           & Advance through planning $\to \cdots \to$ finalization. \\
\bottomrule
\end{tabular}%
\caption{The 17 predefined tools the agent uses to construct and repair
ComfyUI workflows, grouped by function.  Composite macros bundle a
sequence of base graph edits with the correct rewiring and slot binding.
}
\label{tab:tools}
\end{table}

\begin{table}[!htbp]
\centering
\scriptsize
\resizebox{\linewidth}{!}{%
\begin{tabular}{@{}p{0.24\linewidth}p{0.34\linewidth}p{0.33\linewidth}@{}}
\toprule
\textbf{Skill} & \textbf{When to consult (trigger phrases)} & \textbf{Effect on the workflow} \\
\midrule
\texttt{photorealistic}
  & ``photo'', ``photograph'', ``realistic'', ``DSLR'', ``cinematic'',
    ``RAW photo'', ``shot on $\langle$camera$\rangle$''; or ``make it look real''.
  & Rewrites positive/negative prompts toward camera realism; tunes
    KSampler steps, CFG, and sampler/scheduler for natural lighting and
    DSLR-like fidelity. \\
\addlinespace
\texttt{creative}
  & ``creative'', ``artistic'', ``fantasy'', ``concept art'', ``surreal'',
    ``stylized''; clearly fictional scenes where photorealism is wrong.
  & Sets sampler parameters and prompt tags for vivid, concept-art-style
    outputs.  Pairs with \texttt{prompt-artist} for prompt enrichment. \\
\addlinespace
\texttt{high-quality}
  & ``high quality'', ``detailed'', ``sharp'', ``crisp'', ``professional'',
    ``8K''; or verifier reports of soft detail or noise.
  & Layered on top of \texttt{photorealistic}/\texttt{creative}: tunes
    prompt tokens, sampler parameters, and resolution to lift fidelity. \\
\addlinespace
\texttt{prompt-artist}
  & ``aesthetic'', ``dreamy'', ``masterpiece'', ``award-winning'',
    ``imaginative''; or sparse, flat user prompts.
  & Rewrites a flat prompt into vivid multi-dimensional artistic language;
    returns prompt text only and pairs with parameter-tuning skills. \\
\bottomrule
\end{tabular}
}
\caption{The four general-purpose skills loaded into every run before any
benchmark-specific evolution.  Each is a \texttt{SKILL.md} package with
YAML frontmatter exposing only the description at startup; the full body
is loaded on demand via \texttt{read\_skill}.}
\label{tab:skills}
\end{table}

\section{Skill Read Statistics and Workflow Modification Breakdown}
\label{app:skill-read-stats}

Tables~\ref{tab:evolved} and~\ref{tab:modification-breakdown} characterize the
agent's workflow-construction behavior from two complementary angles.
Table~\ref{tab:evolved} measures how much the agent relies on evolved skills
versus the four predefined base skills, broken down by benchmark.
Table~\ref{tab:modification-breakdown} then dissects what kinds of graph edits
the agent actually performs when it constructs or repairs a workflow.

\begin{table}[!htbp]
    \centering
    \small
    \begin{tabular}{lrrrrr}
      \toprule
      Benchmark & Evolved Skills & Total Reads & Base Reads & Evolved Reads & \% Evolved Reads \\
      \midrule
      dpg-bench  &  62 &  9{,}251 &  2{,}078 &  6{,}478 & 70.0\% \\
      geneval2   &  79 & 20{,}282 &  5{,}377 & 11{,}391 & 56.2\% \\
      oneig-en   &  60 &  5{,}355 &  3{,}334 &    872   & 16.3\% \\
      oneig-zh   & 118 &  3{,}266 &  2{,}747 &    244   &  7.5\% \\
      \midrule
      \textbf{Total} & \textbf{318} & \textbf{38{,}154} & \textbf{13{,}536} & \textbf{18{,}985} & \textbf{49.8\%} \\
      \bottomrule
    \end{tabular}
    \caption{Per-benchmark evolved skill read statistics aggregated over
    \textsc{LongCat} and \textsc{Z-Image-Turbo} runs with Claude-Sonnet-4.5.
    \emph{Evolved Skills} counts the number of distinct skills synthesized by the
    skill-evolution loop; \emph{Total Reads} is the aggregate number of
    \texttt{read\_skill} calls across all prompts; \emph{Base Reads} and
    \emph{Evolved Reads} partition those calls by skill origin.
    Benchmarks with richer compositional demands
    (DPG-Bench, GenEval~2) draw far more heavily on evolved skills, whereas the
    OneIG splits---which emphasize single-character stylization---are served
    adequately by the base library.}
    \label{tab:evolved}
\end{table}

Table~\ref{tab:modification-breakdown} further disaggregates the agent's edits
into six semantic event categories.
Prompt-text changes and sampler/guidance adjustments together account for
${\sim}75\%$ of all events across benchmarks, confirming that the agent
primarily operates at the level of conditioning rather than graph topology.
Structural modifications---regional masking, model swaps, and multi-pass
upscaling---are more frequent on GenEval~2, which contains the most demanding
multi-object layout tasks.

\begin{table}[!htbp]
    \centering
    \small
    \setlength{\tabcolsep}{6pt}
    \resizebox{\columnwidth}{!}{%
    \begin{tabular}{l r r r r r r}
    \toprule
    \textbf{Workflow Event} &
    \textbf{DPG-Bench} & \textbf{GenEval2} & \textbf{OneIG-EN} & \textbf{OneIG-ZH} &
    \textbf{Total} & \textbf{\%} \\
    \midrule
    Prompt-text changes              & 2{,}862 & 4{,}139 & 3{,}277 & 3{,}714 & 13{,}992 & 39.3\% \\
    Sampler / guidance hyper-params  & 2{,}776 & 2{,}415 & 3{,}408 & 4{,}161 & 12{,}760 & 35.8\% \\
    Regional / mask graph topology   &    214  & 2{,}958 &    484  &    400  &  4{,}056 & 11.4\% \\
    Model / weight changes (LoRA, ckpt) & 570 &    518  &    721  &    806  &  2{,}615 &  7.3\% \\
    Multi-pass / upscale topology    &    195  &    258  &    328  &    358  &  1{,}139 &  3.2\% \\
    Other (class swaps, removes, misc)& 114    &    599  &    170  &    167  &  1{,}050 &  2.9\% \\
    \midrule
    \textbf{Total}            & \textbf{6{,}731} & \textbf{10{,}887} & \textbf{8{,}388} &
                                       \textbf{9{,}606} & \textbf{35{,}612} & \textbf{100.0\%} \\
    \bottomrule
    \end{tabular}
    }
    \caption{Action-event breakdown for workflow construction in \ours{}.
    Each \emph{event} denotes a node-level workflow change: adding a node,
    removing a node, or editing a node's \texttt{inputs} field.
    Counts are aggregated over the two image-generation models,
    \textsc{LongCat} and \textsc{Z-Image-Turbo}, for each benchmark.}
    \label{tab:modification-breakdown}
\end{table}

\section{Evolved Skill Usage}
\label{app:evolved_skills}
Table~\ref{tab:top10-evolved} provides a usage-level view of the skills produced
by the Claude-Sonnet-4.5.  
Rather than treating these counts as direct measures of
skill quality, we use them to show which reusable recipes the agent actually
consults when solving each benchmark.  The resulting distribution separates
benchmark-specific needs: GenEval~2 emphasizes spatial binding and counting,
DPG-Bench emphasizes material, lighting, and scene detail, and the OneIG
splits emphasize character-count and anime-style control.

\begin{table}[t]
\centering
\scriptsize
\renewcommand{\arraystretch}{1.05}
\setlength{\tabcolsep}{3pt}
\begin{tabular}{@{}c p{0.36\linewidth} r p{0.34\linewidth} r@{}}
\toprule
\textbf{\#}
 & \multicolumn{2}{c}{\textbf{GenEval 2}}
 & \multicolumn{2}{c}{\textbf{DPG-Bench}} \\
\cmidrule(lr){2-3}\cmidrule(l){4-5}
 & \textbf{Skill} & \textbf{Reads}
 & \textbf{Skill} & \textbf{Reads} \\
\midrule
1  & \path|spatial-anchor-with-count|     & 1523 & \path|material-texture-detail|         &  980 \\
2  & \path|spatial-count-binding|         & 1316 & \path|spatial-precision|               &  696 \\
3  & \path|attribute-binding|             &  828 & \path|precise-color-attribution|       &  692 \\
4  & \path|material-modifier|             &  771 & \path|contextual-environment-building| &  591 \\
5  & \path|animal-color-override|         &  633 & \path|lighting-and-reflection-detail|  &  572 \\
6  & \path|exact-seven-count|             &  395 & \path|material-texture-encoding|       &  556 \\
7  & \path|animal-pattern-count|          &  386 & \path|layered-scene-building|          &  348 \\
8  & \path|high-total-count-coordinator|  &  370 & \path|visual-contrast-encoding|        &  297 \\
9  & \path|count-six-five-flock|          & ---  & \path|explicit-colors|                 &  271 \\
10 & \path|spatial-binding-three-objects| & ---  & \path|anatomy-pose-detail|             & ---  \\
\bottomrule
\end{tabular}

\vspace{4pt}

\begin{tabular}{@{}c p{0.36\linewidth} r p{0.34\linewidth} r@{}}
\toprule
\textbf{\#}
 & \multicolumn{2}{c}{\textbf{OneIG-EN}}
 & \multicolumn{2}{c}{\textbf{OneIG-ZH}} \\
\cmidrule(lr){2-3}\cmidrule(l){4-5}
 & \textbf{Skill} & \textbf{Reads}
 & \textbf{Skill} & \textbf{Reads} \\
\midrule
1  & \path|character-counting|                 & 166 & \path|anime-single-character-simple|      & 44  \\
2  & \path|anime-danbooru-ordering|            & 126 & \path|anime-multi-character|              & 33  \\
3  & \path|anime-character-counting|           & 124 & \path|detailed-character-design|          & 33  \\
4  & \path|anime-style-declaration|            &  95 & \path|anime-solo-simple|                  & 21  \\
5  & \path|presentation-slide-text|            &  61 & \path|anime-character-state-verification| & 19  \\
6  & \path|presentation-chart-internal-labels| &  20 & \path|chinese-group-counting|             & 18  \\
7  & \path|presentation-diagram-structure|     &  17 & \path|anime-direct-tag-format|            & 16  \\
8  & \path|rate-limit-mitigation|              &  15 & \path|resolution-quality-tags|            & 15  \\
9  & \path|anime-style-tag|                    & --- & \path|anime-square-resolution|            & --- \\
10 & \path|character-state-control|            & --- & \path|chinese-stylized-prompt|            & --- \\
\bottomrule
\end{tabular}

\caption{Top-10 evolved skills per benchmark for Claude-Sonnet-4.5.  Counts are total \texttt{read\_skill} calls aggregated over all prompts and both image backbones (Z-Image-Turbo and LongCat-Image).  We exclude two run-level aggregator skills so the table reflects topical recipes rather than bookkeeping.}
\label{tab:top10-evolved}
\end{table}
We omit the run-level aggregator skills
\texttt{learned-successes} and \texttt{learned-errors}, which summarize each
cycle's success and failure clusters rather than encoding topical recipes.
Their raw read counts are high by construction: 3135/649/1096/226 and
379/---/53/49 for GenEval~2, DPG-Bench, OneIG-EN, and OneIG-ZH, respectively.

\begin{table}[!htbp]
\centering
\small
\resizebox{\linewidth}{!}{%
\begin{tabular}{@{}p{0.18\linewidth}p{0.30\linewidth}rp{0.26\linewidth}p{0.15\linewidth}@{}}
\toprule
\textbf{Model} & \textbf{LoRA file} & \textbf{Size} & \textbf{Focus} & \textbf{Default strength} \\
\midrule
Z-Image-\newline{}Turbo
  & {\ttfamily Z-Image-Turbo-\newline{}Radiant-Realism-\newline{}Pro.safetensors} & $\sim$163\,MB
  & Realism + lighting polish (``enhancer'')       & 0.80 \\
  & {\ttfamily Z-Image-Turbo-\newline{}Realism-LoRA.\newline{}safetensors}         & $\sim$82\,MB
  & Lighter realism tweak (``realistic snapshot'') & 0.70 \\
\addlinespace
LongCat-Image & \multicolumn{4}{p{0.74\linewidth}@{}}{\textit{--- not supported in our setup ---}} \\
\bottomrule
\end{tabular}
}
\caption{Model-specific LoRA settings used in our experiments.
Z-Image-Turbo is an S3-DiT architecture, so all LoRAs are injected via
\texttt{LoraLoaderModelOnly} (model-only, no CLIP weights).  LongCat-Image
exposes no model tensor compatible with our LoRA tools, so they are
disabled for that backbone.  Strengths follow the recipe in the
\texttt{z-image-turbo} skill.}
\label{tab:loras}
\end{table}

\section{Qualitative Annotations}
\label{app:user_study}
For the user study, each participant was given the instruction template in Box~\ref{box:eval-instructions} and asked to annotate generated images. 
The images were randomly shuffled so that annotators were blind to the corresponding method groups. 
Annotators were paid $\$0.10$ per image, for a total of 2{,}400 annotations and $\$240$ in compensation. 
The annotation task involved no anticipated risk to participants, and our protocol was reviewed and exempted by the Institutional Review Board (IRB).

\begin{tcolorbox}[
  colback=gray!5,
  colframe=gray!60,
  title={\textbf{Evaluation Instructions}},
  fonttitle=\sffamily\bfseries,
  boxrule=0.5pt,
  arc=2pt
]
\label{box:eval-instructions}
Rate the overall result on a \textbf{1--5 scale} that combines three criteria:

\begin{enumerate}[label=(\alph*),leftmargin=2em,itemsep=2pt]
  \item \textbf{Prompt--image alignment} --- Does the image faithfully depict
        the objects, attributes, count, relationships, and style specified in
        the prompt?
  \item \textbf{Visual aesthetic} --- Composition, lighting, colour, and
        balance.
  \item \textbf{Image quality} --- Sharpness, absence of artifacts, anatomical
        and structural correctness, no distorted text/hands/faces.
\end{enumerate}

\vspace{6pt}
\renewcommand{\arraystretch}{1.35}
\begin{tabularx}{\linewidth}{c >{\raggedright\arraybackslash}X}
  \toprule
  \textbf{Score} & \textbf{Description} \\
  \midrule
  1 & \emph{Very poor} --- badly misaligned with the prompt, or severely
      flawed image. \\
  2 & \emph{Poor} --- notable misalignment \textbf{or} notable quality
      issues. \\
  3 & \emph{Average} --- acceptable but with visible alignment \textbf{or}
      quality issues. \\
  4 & \emph{Good} --- well-aligned and visually appealing; only minor
      issues. \\
  5 & \emph{Excellent} --- faithfully depicts the prompt \textbf{and} looks
      polished. \\
  \bottomrule
\end{tabularx}

\end{tcolorbox}

\section{LLM Prompts}
\label{app:prompts}

This section lists the exact prompts used in the implementations of the agent loop
(\S\ref{sec:method:workflow-construction}) and the region-level VLM verifier
(\S\ref{subsec:vlm-verifier}).
All prompts are drawn verbatim from \texttt{comfyclaw/agent.py},
\texttt{comfyclaw/verifier.py}, and \texttt{comfyclaw/harness.py}.

\subsection{Agent System Prompt (\S\ref{sec:method:workflow-construction})}
\label{app:prompts:system}

The following string (\texttt{\_SYSTEM\_PROMPT\_BASE}) is prepended to every
agent conversation as the \texttt{system} role message.
When a model is pinned the \texttt{Pinned image model} paragraph is appended.
The \texttt{<available\_skills>} XML block (skill names and one-line descriptions)
is injected between the base prompt and the pinned-model paragraph.

\begin{lstlisting}
You are ComfyClaw, an expert ComfyUI workflow engineer.  Your job is to BUILD
and GROW ComfyUI workflow topologies -- constructing complete pipelines from
scratch when the workflow is empty, and evolving existing ones in response to
the verifier's region-level feedback.

Iteration strategy
------------------
1. Call report_evolution_strategy first: state your plan and the top issue.
2. Call inspect_workflow to see the current topology.
3. **If the workflow is empty** (no nodes):
   a. Call read_skill("workflow-builder") to load architecture recipes.
   b. Call query_available_models("checkpoints") and query_available_models("diffusion_models")
      to discover available models -- NEVER guess filenames.
   c. Match the model filename to an architecture (SD 1.5, SDXL, Flux, Qwen, etc.)
      using the patterns in the workflow-builder skill.
   d. Build the full pipeline node-by-node using add_node, following the matching recipe.
   e. Use ONLY exact filenames from query results.
   f. Set detailed prompts on the CLIPTextEncode nodes.
   g. Call validate_workflow to catch wiring errors before submitting.
   h. Call finalize_workflow (it auto-validates and blocks if errors remain).
4. **If the workflow already has nodes**, follow the evolution strategy:
   a. Call set_prompt -- craft a detailed, professional positive prompt AND a strong
      negative prompt based on the user's goal (see "Prompt engineering" below).
      Do this EVERY iteration, even if you also plan structural changes.
   b. If a relevant skill is listed in <available_skills>, call read_skill to load
      its full instructions BEFORE applying that upgrade.
   c. Call query_available_models BEFORE adding any LoRA node.
   d. Apply structural upgrades (LoRA / regional / hires / inpaint).
   e. Tune sampler parameters (steps, CFG, seed) as needed.
   f. Call validate_workflow to catch wiring errors.
   g. Call finalize_workflow when done (it auto-validates).

Prompt engineering (step 3)
----------------------------
The workflow's positive prompt is pre-seeded with the user's raw goal text.
You MUST replace it with a professional-quality prompt every iteration.

Positive prompt -- structure:
  [subject & scene], [style], [lighting], [camera/lens], [quality boosters]
  * Expand every meaningful concept: vague nouns -> vivid adjectives + nouns.
  * Add artistic / photographic style: "cinematic", "concept art", "photorealistic",
    "watercolor painting", "isometric", etc.
  * Add lighting: "golden hour", "dramatic rim lighting", "neon glow", "soft diffuse".
  * Add quality boosters: "8k", "ultra detailed", "sharp focus", "ray tracing", "award winning".
  * If the image has multiple subjects/regions, describe each clearly.

Negative prompt -- always include these baseline entries, then add scene-specific ones:
  "blurry, out of focus, low quality, low resolution, noisy, grainy, jpeg artifacts,
   watermark, text, signature, ugly, bad anatomy, deformed, disfigured,
   poorly drawn hands, extra fingers, mutated limbs, cloned face, plastic skin"

Example (input: "a cyberpunk city at night"):
  positive: "a futuristic cyberpunk city skyline at night, towering neon-lit skyscrapers,
   wet reflective streets, holographic advertisements, dense rain, cinematic composition,
   dramatic volumetric lighting, wide angle lens 24mm, 8k, photorealistic, ultra detailed,
   sharp focus, ray tracing, blade runner aesthetic"
  negative: "blurry, low quality, noisy, watermark, text, bad anatomy, deformed, ugly,
   cartoon, anime, daytime, sunny, empty street"

Structural upgrade priority (iteration 2+)
------------------------------------------
When the workflow already has nodes AND verifier feedback is present:
  * Do NOT just refine the prompt -- prompt-only changes plateau quickly.
  * PREFER structural upgrades: LoRA, hires-fix, regional, inpaint.
  * If ANY region_issue has fix_strategies containing "inject_lora_*",
    you MUST attempt that structural upgrade, not fall back to prompt tweaking.
  * Combine: always refine the prompt AND add a structural upgrade together.
  * Only fall back to prompt-only when no LoRA / inpaint models are installed
    or the fix strategies are exclusively prompt-related.

Human-in-the-loop feedback
--------------------------
When the verifier feedback section starts with "## Human Reviewer Feedback",
the feedback comes from a human reviewer, not an automated VLM.
Human feedback expresses subjective preferences -- style, mood, composition,
color palette, artistic direction.  Prioritize these over structural/technical
changes.  Items prefixed with [HUMAN] are direct human requests -- address
each one specifically.  Do not second-guess or override human preferences.

[Pinned image model paragraph -- appended when a model is locked]
The image-generation model for this session is LOCKED to: {model_name}
  * DO NOT change the ckpt_name / unet_name of any loader node.
  * You MAY add LoRA loaders on top of the pinned model.
  * If the server has no models available (offline / dry-run), skip any action
    that requires model discovery and focus on prompt / sampler tuning.
\end{lstlisting}

\subsection{Agent User Message Template (\S\ref{sec:method:workflow-construction})}
\label{app:prompts:user}

Each agent invocation receives a dynamically assembled \texttt{user} message
(built by \texttt{\_build\_user\_message} in \texttt{agent.py}).
The template below shows all sections that may appear; sections are joined
with double newlines and omitted when their data are absent.

\begin{lstlisting}
## Image Goal (user's original request)
{original_prompt}

## Iteration
{iteration}

## Current Positive Prompt (baseline -- needs refinement)
{current_positive_prompt_from_workflow}
Use `set_prompt` to replace this with a detailed, high-quality version.

## Active Model
`{checkpoint_or_unet_filename}` ({architecture description from arch.yaml})

## Pre-loaded Skill: {model_skill_name}          [when a model-specific skill exists]
{full skill body -- agent does NOT need to call read_skill for this}

## Learned Skills (from past experience on this model)  [when evolved skills exist]
{description of evolved skill}  ->  read_skill("{evolved_skill_name}")

## Suggested Skills                               [when keyword-matched built-in skills exist]
These skills may be relevant: {comma-separated skill names}
Call read_skill(<name>) to load full instructions before applying.

## Verifier Feedback (previous iteration)         [from iteration 2 onward]
{VerifierResult.format_feedback() output}

REQUIRED structural upgrades (from verifier fix_strategies):
  * `inject_lora_detail`  ->  read_skill("lora-enhancement") then add_lora_loader
  * `add_hires_fix`       ->  read_skill("hires-fix") then add_hires_fix
  ...

## Human Reviewer Feedback (previous iteration)   [hybrid / human verifier mode]
{human feedback text with [HUMAN] prefixed items}
Prioritize their subjective preferences (style, mood, composition, color).

## Memory / Past Attempts
Attempt {n} (score={s:.2f}):
  Passed: {passed requirements}
  Failed: {failed requirements}
  Experience: {one-line lesson from that attempt}

## CRITICAL -- Empty Workflow -- You MUST Build From Scratch   [when workflow is empty]
The workflow is COMPLETELY EMPTY.  You CANNOT finalize without adding nodes.
Step 1: Call `read_skill("workflow-builder")` ...
Step 2: Call `query_available_models('checkpoints')` ...
...

Begin with report_evolution_strategy, then inspect_workflow,
apply your changes, then finalize_workflow.
\end{lstlisting}

\subsection{Workflow Repair Prompt (\S\ref{sec:method:workflow-construction})}
\label{app:prompts:repair}

When ComfyUI rejects a workflow submission (HTTP~4xx, validation failure, or
execution error), the harness re-invokes the agent with the following feedback
string (\texttt{\_build\_repair\_feedback} in \texttt{harness.py}) as the
\texttt{verifier\_feedback} argument.

\begin{lstlisting}
## ComfyUI Rejected the Workflow -- Repair Required

Your last workflow submission was rejected with the following error:
```
{verbatim ComfyUI error string}
```

**Repair protocol (follow in order):**
1. Call `inspect_workflow` to see the FULL current topology and all connections.
2. Call `validate_workflow` to get a list of graph errors (dangling refs, wrong slots).
3. For each error:
   - If a node references a nonexistent source -> fix with `connect_nodes` or `delete_node`
   - If a slot index is wrong -> `delete_node` the broken node and `add_node` a new one
     with correct wiring
   - If a model/filename is wrong -> use `query_available_models` to get exact names,
     then `set_param`
   - If a node class doesn't exist -> `delete_node` it and use a different class_type
4. Call `validate_workflow` again to confirm all issues are resolved.
5. Call `finalize_workflow` (it will auto-validate and block if still broken).

**IMPORTANT:** Do NOT just add new nodes on top of broken ones -- `delete_node` the
broken node first, then `add_node` a replacement with correct connections.

**Output slot reference:**
  CheckpointLoaderSimple -> slot 0: MODEL, slot 1: CLIP, slot 2: VAE
  UNETLoader / CLIPLoader / VAELoader -> slot 0 only
  KSampler -> slot 0: LATENT
  VAEDecode -> slot 0: IMAGE
  CLIPTextEncode -> slot 0: CONDITIONING

[When a previous VerifierResult is available:]
-- Previous Verifier Feedback (for context) --
{VerifierResult.format_feedback() output}
\end{lstlisting}

\subsection{Verifier Prompts (\S\ref{subsec:vlm-verifier})}
\label{app:prompts:verifier}

The verifier (\texttt{comfyclaw/verifier.py}) uses three distinct prompts
corresponding to the two-pass pipeline described in \S\ref{subsec:vlm-verifier}.

\paragraph{Pass~1 -- Requirement decomposition (\texttt{\_DECOMPOSE\_PROMPT}).}
A text-only call (no image) that breaks the user prompt into a list of
yes/no questions.  In the default batched mode (\texttt{batch\_mode=True})
this step runs once per unique prompt and is cached for all subsequent
iterations, so each unique prompt pays the decomposition cost only once.

\begin{lstlisting}
Analyze the following image generation prompt and break it down into specific,
observable visual requirements. For each, write a yes/no question answerable from the image.

Respond ONLY with a JSON array of question strings.

Prompt: {prompt}
\end{lstlisting}

\paragraph{Pass~2 -- Batched unified verification (\texttt{\_UNIFIED\_VERIFY\_PROMPT}).}
The default path.  A single vision call answers all decomposed yes/no
questions and simultaneously produces region-level issues, evolution
suggestions, and a holistic 1--10 score.  The image is uploaded exactly
once per verify call.

\begin{lstlisting}
You are an expert image quality analyst and ComfyUI workflow engineer.

You are given one generated image and the intended prompt. Do BOTH tasks in a
SINGLE pass and return ONE JSON object. Do not emit any prose, markdown, or
code fences -- just the JSON.

TASK 1 -- Requirement checks: Answer each yes/no question below based on the
image. Use strict "yes" / "no" answers (lower-case). Every question MUST be
answered; never skip.

Questions (answer all):
{questions_block}

TASK 2 -- Holistic analysis: Produce a short overall assessment, an integer
score (1-10), a list of region-level issues, and a list of concrete workflow
evolution suggestions.

Score rubric (integer 1-10):
  1-2: Completely wrong -- unrecognizable, no relation to prompt
  3-4: Major failures -- wrong subject, severe artifacts, missing key elements
  5-6: Partial match -- right subject but significant quality or accuracy issues
  7-8: Good -- matches prompt well with minor issues (slight artifacts, soft details)
  9-10: Excellent -- faithful to prompt, high quality, minimal or no issues

Fix strategy vocabulary (use these exact strings):
  inject_lora_detail   | inject_lora_style    | inject_lora_anatomy  | inject_lora_lighting
  add_regional_prompt  | add_hires_fix        | add_inpaint_pass     | add_ip_adapter
  refine_positive_prompt | refine_negative_prompt | increase_steps | adjust_cfg | adjust_sampler

Return exactly this JSON schema:
{
  "requirements": [
    {"question": "<verbatim question text>", "answer": "yes" or "no"}
  ],
  "overall_assessment": "<1-2 sentence overall quality summary>",
  "score": <integer 1-10>,
  "region_issues": [
    {
      "region": "<foreground subject | background | face | hands | sky | ...>",
      "issue_type": "<anatomy | texture | lighting | artifact | composition | detail | color | proportion>",
      "description": "<specific problem description>",
      "severity": "<low | medium | high>",
      "fix_strategies": ["<workflow action 1>", "<workflow action 2>"]
    }
  ],
  "evolution_suggestions": [
    "<concrete workflow change 1>",
    "<concrete workflow change 2>"
  ]
}

CRITICAL: requirements MUST contain exactly {n_questions} entries, in the same
order as the questions above.

Intended prompt: {prompt}
\end{lstlisting}

\paragraph{Pass~2 -- Legacy detailed analysis (\texttt{\_DETAILED\_ANALYSIS\_PROMPT}).}
Used as a fallback when the unified call fails to parse, or when
\texttt{batch\_mode=False}.  This call receives the image and returns the
holistic analysis only; requirement checks are handled by separate per-question
calls in that mode.

\begin{lstlisting}
You are an expert image quality analyst and ComfyUI workflow engineer.

Analyze this generated image against the intended prompt, then return a JSON object with:
{
  "overall_assessment": "<1-2 sentence overall quality summary>",
  "score": <integer 1-10>,
  "region_issues": [
    {
      "region": "<specific area: foreground subject | background | face | hands | sky | ...>",
      "issue_type": "<anatomy | texture | lighting | artifact | composition | detail | color | proportion>",
      "description": "<specific problem description>",
      "severity": "<low | medium | high>",
      "fix_strategies": ["<workflow action 1>", "<workflow action 2>"]
    }
  ],
  "evolution_suggestions": [
    "<concrete workflow change 1: what to add/modify and why>",
    "<concrete workflow change 2>"
  ]
}

Score rubric (integer 1-10):
  1-2: Completely wrong -- unrecognizable, no relation to prompt
  3-4: Major failures -- wrong subject, severe artifacts, missing key elements
  5-6: Partial match -- right subject but significant quality or accuracy issues
  7-8: Good -- matches prompt well with minor issues (slight artifacts, soft details)
  9-10: Excellent -- faithful to prompt, high quality, minimal or no issues

Fix strategy vocabulary (use these exact strings):
  inject_lora_detail   | inject_lora_style    | inject_lora_anatomy  | inject_lora_lighting
  add_regional_prompt  | add_hires_fix        | add_inpaint_pass     | add_ip_adapter
  refine_positive_prompt | refine_negative_prompt | increase_steps | adjust_cfg | adjust_sampler

Intended prompt: {prompt}
\end{lstlisting}


\end{document}